\documentclass[10pt,twocolumn,letterpaper]{article}
\usepackage[linesnumbered,ruled,vlined]{algorithm2e}
\usepackage{cvpr}      
\usepackage{algorithmic}
\usepackage{multirow}
\usepackage{pifont}
\usepackage{caption}
\usepackage{float}
\usepackage{arydshln}

\usepackage[table]{xcolor}
\usepackage{colortbl}

\usepackage{booktabs}
\usepackage{array}

\usepackage[dvipsnames]{xcolor}

\usepackage{bbding}
\definecolor{cvprblue}{rgb}{0.21,0.49,0.74}
\usepackage[pagebackref,breaklinks,colorlinks,allcolors=cvprblue]{hyperref}

\def\paperID{19515} 
\def\confName{CVPR}
\def\confYear{2026}

\title{
\textit{GraspALL}: Adaptive Structural Compensation from Illumination Variation for Robotic Garment Grasping in Any Low-Light Conditions
} 

\author{Haifeng Zhong$^{1}$, Wenshuo Han$^{1}$, Zhouyu Wang$^{1}$, Runyang Feng$^{1}$, Fan Tang$^{2}$, Tong-Yee Lee$^{3}$, \\Zipei Fan$^{1}$, Ruihai Wu$^{4}$, Yuran Wang$^{4}$, Hao Dong$^{4}$, Hechang Chen$^{1,6}$,\\
Hyung Jin Chang$^{5}$, Yixing Gao$^{1,6\ast}$\\
$^{1}$ School of Artificial Intelligence, Jilin University,
 $^{2}$ Chinese Academy of Sciences,\\
$^{3}$ National Cheng-Kung University,
$^{4}$ Peking University,
$^{5}$ University of Birmingham\\
$^{6}$ Engineering Research Center of Knowledge-Driven Human-Machine Intelligence, MoE, China,\\
{\tt\small zhonghf23@mails.jlu.edu.cn, gaoyixing@jlu.edu.cn}
}

\begin{document}
\twocolumn[{
\renewcommand\twocolumn[1][]{}
\maketitle
\begin{center}
\vspace{-0.9cm}
\includegraphics[width=\textwidth]{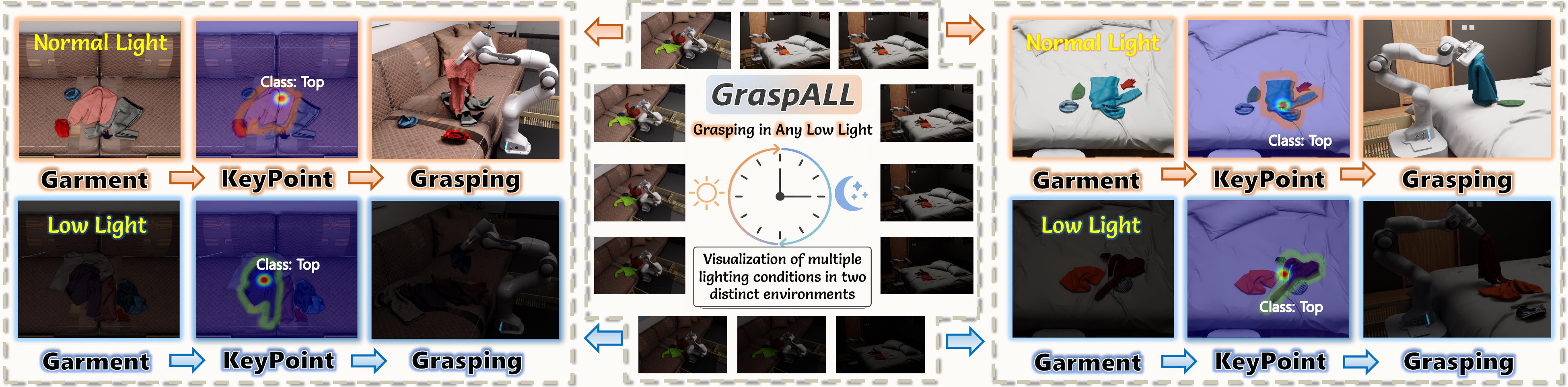}
\vspace{-0.7cm}
\captionof{figure}{
The grasping performance of our GraspALL in the living room (left) and bedroom (right) scenes under different illumination conditions.
Given the variability of household scenes and lighting, service robots should possess all-day garment perception capabilities.
}
\label{fig:1}
\vspace{-0.2cm}
\end{center}
}]

\newcommand\blfootnote[1]{%
  \begingroup
  \setlength{\footnotesep}{0pt} 
  \renewcommand\thefootnote{}\footnote{\vspace{-0cm}#1}
  \addtocounter{footnote}{-1}%
  \endgroup
}
\blfootnote{$^\ast$ Corresponding author}

\begin{abstract}
Achieving accurate garment grasping under dynamically changing illumination is crucial for all-day operation of service robots. 
However, the reduced illumination in low-light scenes severely degrades garment structural features, leading to a significant drop in grasping robustness. 
Existing methods typically enhance RGB features by exploiting the illumination-invariant properties of non-RGB modalities, yet they overlook the varying dependence on non-RGB features under varying lighting conditions, which can introduce misaligned non-RGB cues and thereby weaken the model's adaptability to illumination changes when utilizing multimodal information.
To address this problem, we propose GraspALL, an illumination-structure interactive compensation model.  
The innovation of GraspALL lies in encoding continuous illumination changes into quantitative references to guide adaptive feature fusion between RGB and non-RGB modalities according to varying lighting intensities, thereby generating illumination-consistent grasping representations.
Experiments on the self-built garment grasping dataset demonstrate that GraspALL improves grasping accuracy by 32-44\% over baselines under diverse illumination conditions.
The code is available at \href{https://github.com/Zhonghaifeng6/GraspALL}
{https://github.com/Zhonghaifeng6/GraspALL}
\end{abstract}    
\section{Introduction}
Garment grasping is a fundamental capability for service robots in daily tasks such as cleaning~\cite{R10,R11} and dressing assistance~\cite{R12,R13,R14}. 
While existing methods~\cite{R10,R12,R46,R47} achieve high grasping accuracy under normal illumination, lighting conditions in real household environments are often dynamic. 
For instance, in patients, elderly and infant care scenarios, robots are frequently required to operate in low-light or even unlit environments to avoid disturbance. 
As shown in Fig.~\ref{fig:1}, this necessitates that robots possess all-day perceptual capability, ensuring reliable performance under any illumination.
However, reduced illumination severely degrades garment texture, wrinkles, and edge details, thereby diminishing the robustness of grasping.

\begin{figure}[t]
  \centering
  \includegraphics[height=3.15cm]{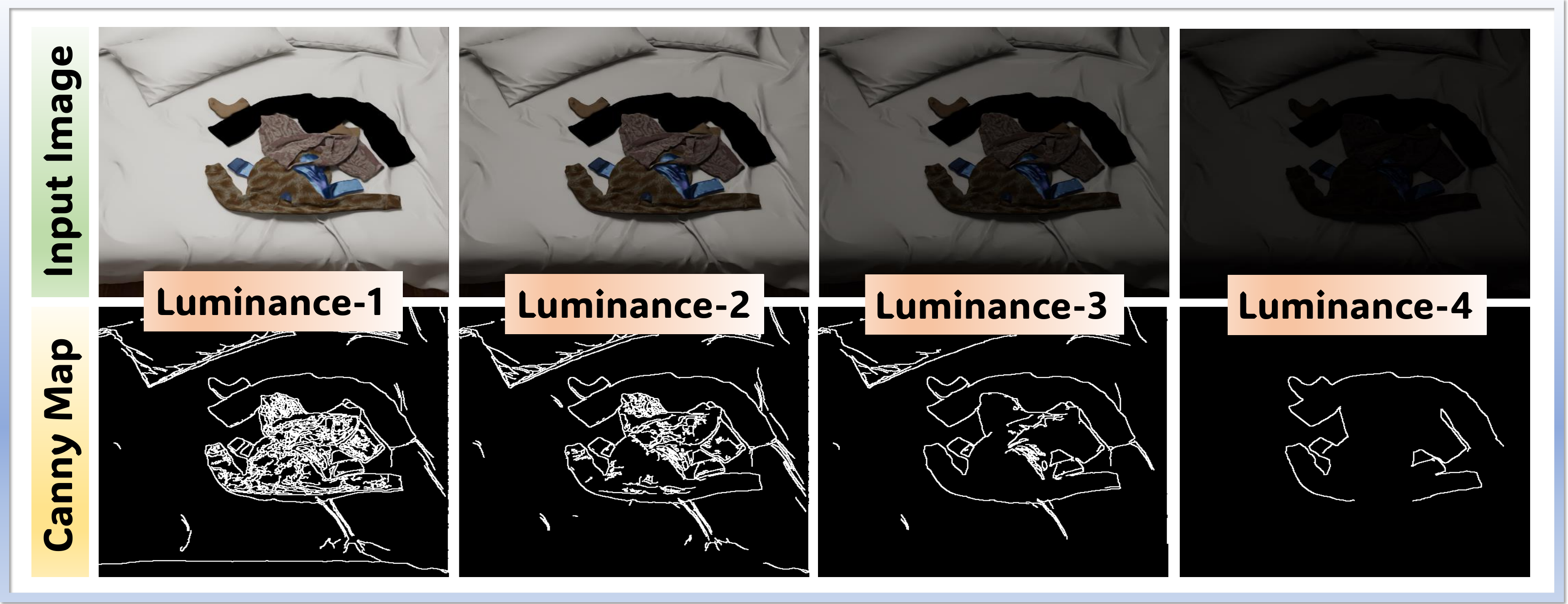}
  \vspace{-0.3cm}
  \caption{
   Structure maps generated by the Canny operator from identical scenes under different luminance levels.
  }
  \vspace{-0.7cm}
  \label{fig:f2}
\end{figure}

To address the perceptual degradation caused by illumination variations, existing methods~\cite{R1,R2,R3,R19} generally adopt multimodal fusion~\cite{R15,R16,R18} to enhance RGB features by leveraging the structural illumination invariance of non-RGB modalities (e.g., depth map). 
Although the above methods are effective, they often overlook the differences in the model's requirements for structural features of non-RGB under varying illumination conditions.
As shown in Fig.~\ref{fig:f2}, illumination changes can significantly distort the geometry of garments, resulting in inconsistent structural cues for the same garment. 
Such illumination-induced discrepancies cause the feature response of RGB under illumination variations are suppressed by the stronger structural signals of non-RGB, leading the model to prioritize non-RGB cues over subtle yet critical RGB luminance information, thereby reducing robustness to illumination changes.

Based on the above analysis, we hold that a more rational paradigm is to enable the model to perceive the input illumination levels and then extract appropriate structural compensation from non-RGB modalities according to different illumination levels, thereby conditionally enhancing garment representations~\cite{R17}. 
This paradigm facilitates the collaborative fusion of cross-modal features to cope with illumination variations, while we also identify two challenges it poses: 
\textbf{(1)} how to accurately estimate the input illumination level to provide quantitative guidance for cross-modal feature fusion; 
\textbf{(2)} based on the estimation of illumination levels, how to induce non-RGB to generate structural compensation adaptive to illumination changes.

\begin{figure*}[t]
  \centering
  \includegraphics[height=8.4cm]{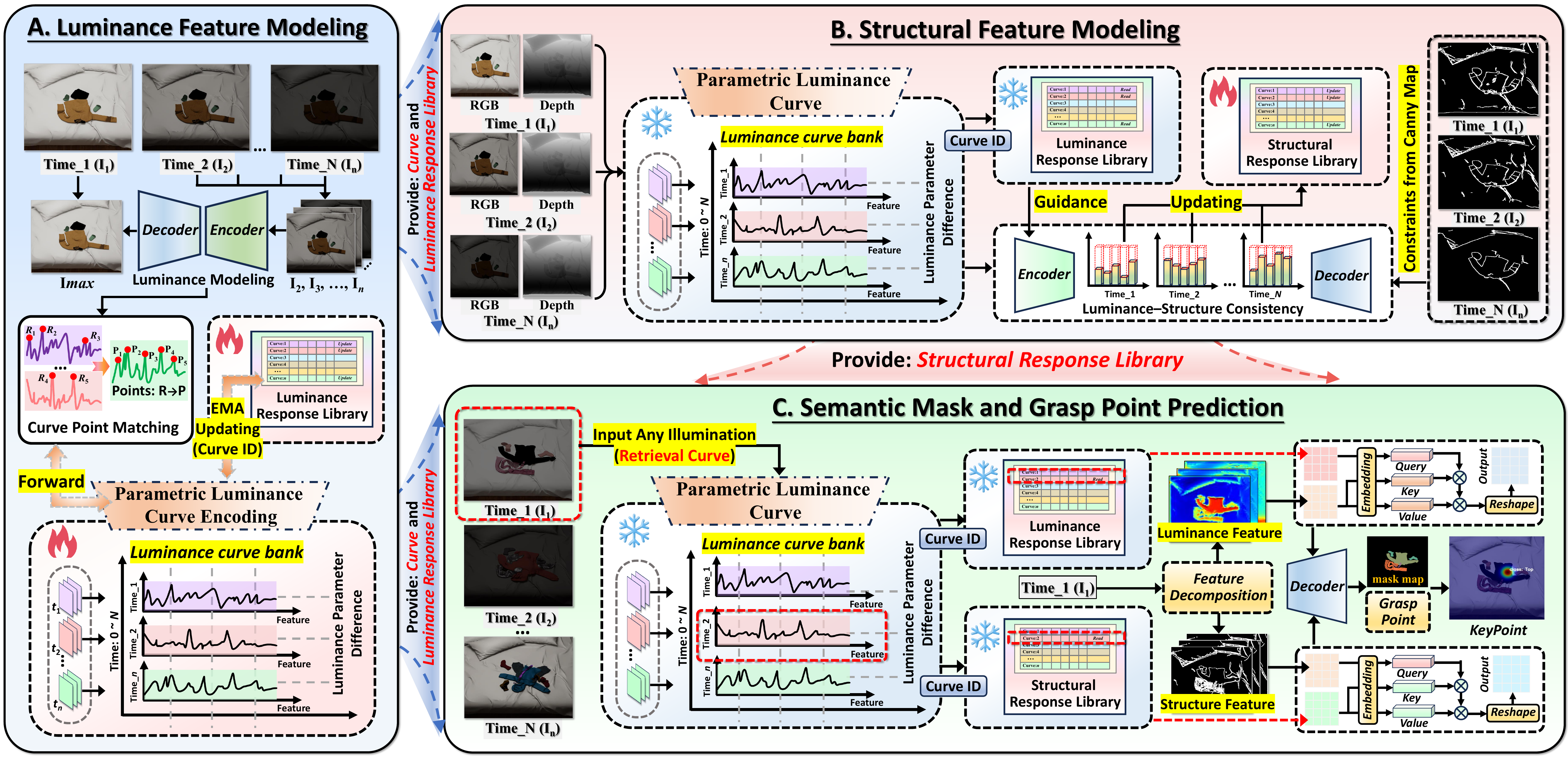}
  \vspace{-0.25cm}
  \caption{
   Overview of our GraspALL. 
 (A) describes the generation process of the Parametric Luminance Curve and luminance compensation features (in Sec.~\ref{s31}).
(B) illustrates the generation process of structural compensation features adaptive to the input luminance (in Sec.~\ref{s32}).
(C) outlines the generation process of grasp points based on the luminance and structural compensation features (in Sec.~\ref{s33}).
  }
  \vspace{-0.5cm}
  \label{fig:f3}
\end{figure*}

To address the above challenges, as shown in Fig.~\ref{fig:f3}, we propose a novel model capable of adapting to illumination variations to enable garment \underline{\textbf{Grasp}}ing in \underline{\textbf{A}}ny \underline{\textbf{L}}ow-\underline{\textbf{L}}ight (named \textbf{GraspALL}).
Unlike methods~\cite{R1,R2,R3,R19} that treat non-RGB as static supplements, the innovation of our GraspALL lies in pioneering a parametric luminance representation method and an illumination-adaptive structural compensation strategy to guide non-RGB in adaptively enhancing garment features in RGB according to different illumination levels. 
Specifically, we first propose a parametric luminance curve (PLC), which fits representative luminance patterns of inputs under different illumination via learnable parameter sets, enabling general representation of any illumination level. 
Based on the input luminance estimated by the PLC, we derive the required luminance compensation features during the luminance restoration process to drive depth maps to generate corresponding structural compensation features.
We then calculate the feature correlation scores between depth maps and luminance compensation features to suppress the weights of features incompatible with the current illumination level, thus obtaining illumination-adaptive structural compensation features for supporting grasp point modeling process.

Moreover, considering the scarcity of multi-illumination garment grasping datasets, we construct a dataset for grasping tasks under diverse illumination conditions.     
Unlike previous works~\cite{R7,R40,R41}, the novelty of our dataset lies in that it consists of diverse garment categories, covers typical layouts including sofa and bed furniture, and incorporates diverse illumination variations from bright to dim—thus simulating household scenes with dynamic illumination changes.
Experiments conducted under different illumination levels show that compared with baselines, our GraspALL can improve grasping accuracy by 32\%–44\%.

The contributions of this work can be summarized as:
\begin{itemize}
\item[$\bullet$]
We present the first systematic analysis of how dynamic illumination variations affect garment grasping, uncovering critical challenges overlooked by existing methods and offering new insights for designing illumination-robust garment grasping models.
\end{itemize}

\begin{itemize}
\item[$\bullet$]
We propose a GraspALL model for garment grasping in varying illumination, which can guide non-RGB to conditionally enhance RGB garment features from a illumination adaptation perspective.
\end{itemize}

\begin{itemize}
\item[$\bullet$]
We introduce a new task of garment classification and grasping under illumination variation, and establish a benchmark comprising a large-scale dataset and diverse household scenarios, providing a unified testbed for evaluating garment grasping under illumination changes.
\end{itemize}

\section{Related work}
\textbf{Garment Grasping.} 
Garment grasping has broad applications in household scenarios. 
Existing methods commonly rely on object detection~\cite{R20,R21,R27}, relation detection~\cite{R20,R24,R25,R26}, or learning-based strategies~\cite{R21,R22,R48,R53} to achieve grasping; some further incorporate semantic segmentation~\cite{R23,R21,R28,R29} to enhance perception of garments. 
While the above methods have achieved notable progress, they generally assume stable image quality and thus struggle to cope with the dynamic illumination variations that naturally occur across time and space in household scenes. 
When the illumination changes from light to dark, garment features often undergo severe degradation, leading to a significant drop in the robustness of grasp point prediction. 
In contrast, our work provides a detailed analysis of the challenges posed by illumination variation for garment grasping, and addresses these challenges by reinforcing cross-modal features to capture critical garment representations, thereby enabling more accurate grasping.

\noindent\textbf{Garment Grasping of Illumination Variations.}
Faced with dynamically changing illumination, existing methods~\cite{R7,R30,R31,R32,R33} focus on incorporating non-RGB modalities such as depth maps, exploiting their illumination-invariant structural properties to complement RGB, thereby improving grasping performance under low light. 
Although the above methods are partially effective, they largely overlook how to achieve dynamic complementarity between RGB and non-RGB features under illumination variation. 
Since illumination changes dynamically affect garment structural characteristics, the model’s reliance on non-RGB structural features should vary across different lighting conditions. 
In contrast, our work is the first to systematically address the impact of illumination variation on garment grasping, and the proposed GraspALL enables grasping across arbitrary illumination conditions by guiding depth maps to adapt to illumination-caused degradation.
\section{Method}
To tackle the challenges of illumination variation in garment grasping, as shown in Fig.~\ref{fig:f3}, we propose GraspALL, a grasp point recognition model built on luminance–structure interactive compensation.  
GraspALL consists of three core components: the parametric luminance curve, the luminance response library, and the structural response library. 
\subsection{Luminance Feature Modeling}
\label{s31}
Traditional luminance estimation methods~\cite{R35,R36} typically rely on histograms~\cite{R17,R51}, but the non-learnable nature of histograms makes it difficult to adapt to diverse illumination variations.
To address this, we present a learnable Parametric Luminance Curve (PLC), whose distinctiveness lies in adopting a learnable parameter set to uniformly represent the representative luminance patterns across various illuminations, thereby generating more robust luminance interpretations.
The performance of the PLC is verified in Sec.\ref{46}.

Firstly, we need to define a luminance curve bank $\mathbf{C}=\{\mathbf{C}_{1},\mathbf{C}_{2},...,\mathbf{C}_{N}\},N=12$, each curve $\mathbf{C}_{n} \in \mathbb{R}^{R}$ consists of \textit{R} discrete sampling points parameterized by learnable raw parameters $\mathbf{P}_{n}=\{\mathbf{P}_{n,1},\mathbf{P}_{n,2},...,\mathbf{P}_{n,R}\},R=256$. 
$\mathbf{P}_{n}$ is a learnable parameter that enables the model to adaptively learn the optimal brightness curve of different illumination conditions.
The analysis of \textit{N} and ${R}$ in luminance curve bank $\mathbf{C}$ is provided in the supplementary materials.

Next, we need to define a learning objective for the luminance curve bank $\mathbf{C}$. 
Given a set of images $\{\mathbf{I}_{1},\mathbf{I}_{2},...,\mathbf{I}_{N}\}$ under varying illumination, we select the image $\mathbf{I}_{max}$ with the highest luminance based on histogram statistics as the reference luminance image. 
For the other images $\mathbf{I}_{n}$ except $\mathbf{I}_{max}$, we compute ${R}$ representative luminance values using histograms with ${R}$ intervals to form a set $\mathbf{H}_R$, and then identify the curve that has the most matching points with $\mathbf{H}_R$, obtaining the corresponding curve index $\mathbf{ID}_{n}$:
\begin{equation}
\begin{array}{l}
\mathbf{ID}_{n} = \mathbf{argmin}\left|| \mathbf{H}_{i}-\mathbf{C}\left(\mathbf{P}_{n,i}\right)\right||, i \in R, n \in N.\\
\end{array}
\label{eq2}
\end{equation}

We take the reference luminance image $\mathbf{I}_{max}$ as the luminance anchor and align the luminance features of other images $\mathbf{I}_{n}$ to $\mathbf{I}_{max}$ through a shared encoder–decoder:
\begin{equation}
\begin{array}{l}
\mathbf{I}_{max}^{n} = \mathcal{D}\left( \mathcal{E}\left(\mathbf{I}_{n}\right)\right)\leftarrow \mathcal{L}_{sc}\left(\mathbf{I}_{max}-\mathbf{I}_{max}^{n}\right)_{\textbf{L1}},\\
\end{array}
\end{equation}
where $\mathcal{E}\left(\cdot\right)$ and $\mathcal{D}\left(\cdot\right)$ denote the encoder and decoder~\cite{R34}, and encoder features: $\mathbf{F}_{en}^{n}=\mathcal{E}\left(\mathbf{I}_n\right)$. ``$\leftarrow$'' represents the loss supervision. $\mathcal{L}_{sc}$ represents spectral consistency loss.

By aligning features with the brightest sample, the model is able to gain the luminance compensation features essential for restoring the luminance of inputs.
These features not only enhance garment distinguishability but also implicitly reflect structural deficiencies caused by illumination variations, allowing them to further guide depth maps to generate corresponding structural compensation features.
To this end, we construct a Luminance Response Library $\mathbf{M}_{L}$, to store luminance features corresponding to different illumination conditions: $\mathbf{M}_{L}=\{\mathbf{M}^{1}_{L},\mathbf{M}^{2}_{L},...,\mathbf{M}^{N}_{L}\}$.
\begin{figure}[t]
  \centering
  \includegraphics[height=3cm]{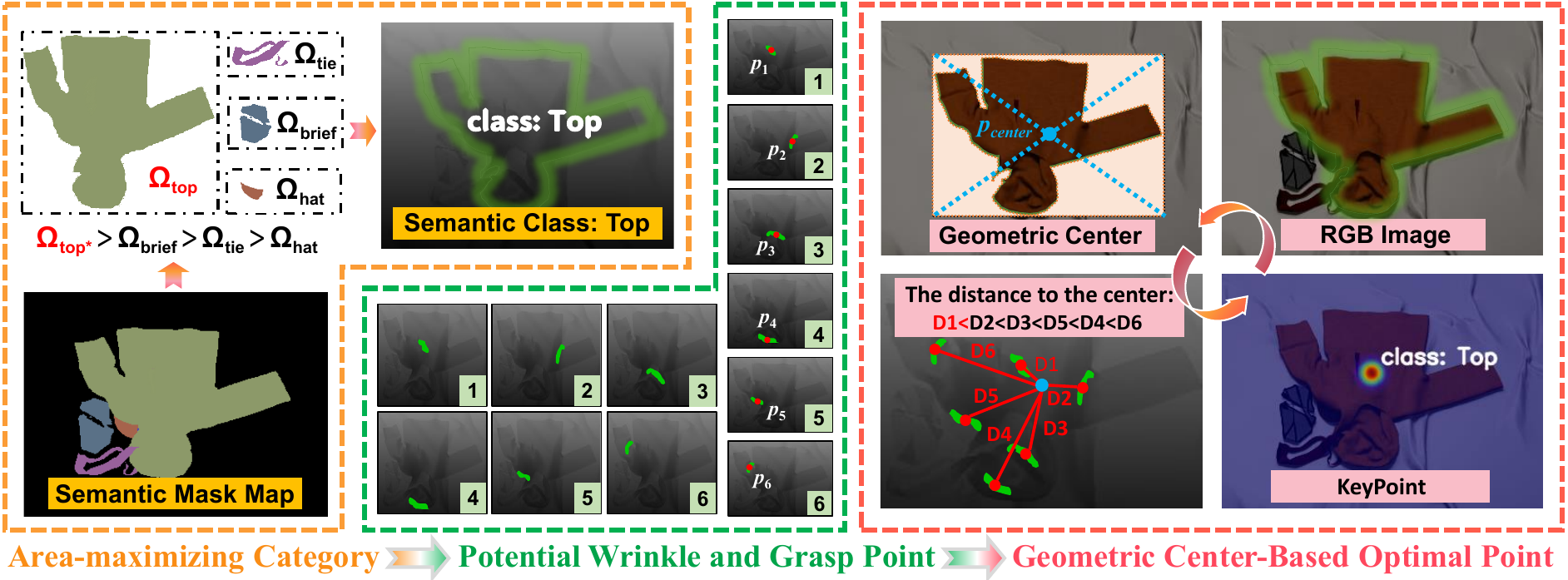}
  \vspace{-0.3cm}
  \caption{
Flowchart of proposed depth-optimal search strategy.
  }
  \vspace{-0.6cm}
  \label{fig:f8}
\end{figure}

Based on the index $\mathbf{ID}_{n}$, the encoder feature $\mathbf{F}_{en}^{n}$ of the alignment process is updated into the corresponding slot of $\mathbf{M}_{L}$ using an exponential moving average (EMA)~\cite{R39}:
\begin{equation}
\begin{array}{l}
\displaystyle
\mathbf{M}_{L}=\left(1-\alpha\right)\mathbf{M}_{L}^{n}+\alpha\cdot\mathbf{F}_{en}^{n},n=\mathbf{ID}_{n},\\
\end{array}
\label{eq4}
\end{equation}
where $\alpha=0.05$ is the EMA momentum. 
The analysis of the $\alpha$ is provided in the supplementary materials.

To enable the adaptive learning of the luminance curve bank $\mathbf{C}$, we introduce a spectral consistency loss $\mathcal{L}_{{sc}}$~\cite{R38}. 
$\mathcal{L}_{sc}$ emphasizes clustering consistency by minimizing the L1 distance between the luminance features $\mathbf{F}_{en}^n$ of the current input and the corresponding slot features in $\mathbf{M}_L$, thereby encouraging the curve bank to dynamically adjust its parameters for more accurate luminance indexing. 
By minimizing $\mathcal{L}_{{sc}}$, the network optimizes the selected curve parameters $\mathbf{P}_{n}$ through chained gradient backpropagation:
\begin{equation}
\begin{array}{l}
\displaystyle
\frac{\partial\left(\mathcal{L}_{sc}\right)}{\partial\left(\mathbf{P}_{n}\right)}=\frac{\partial\left(\mathcal{L}_{sc}\right)}{\partial\left(\mathbf{M}_L\right)}\cdot\frac{\partial\left(\mathbf{M}_L\right)}{\partial\left(\mathbf{F}_{en}^{n}\right)}\cdot\frac{\partial\left(\mathbf{F}_{en}^{n}\right)}{\partial\left(\mathbf{C}\right)}\cdot\frac{\partial\left(\mathbf{C}\right)}{\partial\left(\mathbf{P}_{n}\right)},\\
\end{array}
\label{ee4}
\end{equation}
where $\partial \left(\cdot\right)$ denotes the gradient of the current parameters with respect to the output parameters of the previous layer.

Eqs.\ref{eq2}–\ref{ee4} uniquely emphasize the learnability of illumination interpretation, and they ultimately generate a luminance curve bank and the luminance response library that are used to guide the generation of the structural response library detailed in Sec.\ref{s32}.
Although real-illumination involves multiple sources, reflections, and material effects, PLC is primarily designed to capture single and important illumination factor that influence garment appearance, and other factors will be investigated in the future work.
\subsection{Structural Feature Modeling}
\label{s32}
Although structurally stable in extreme low light, depth maps lack discriminative appearance details. Because highly deformable garments yield similar geometric structures, relying solely on depth causes class confusion. Conversely, degraded RGB images still preserve complementary semantic cues (detailed analysis in Suppl. Sec.1). 
However, existing methods~\cite{R1,R2} are inherently static and struggle with dynamic illumination. To address this, our structural modeling conditionally extracts adaptive depth features guided by explicit illumination understanding.

Given a set of images $\{\mathbf{I}_{1},\mathbf{I}_{2},...,\mathbf{I}_{N}\}$ ranging from dark to bright, for one of the images $\mathbf{I}_n$, we employ Eq.~\ref{eq2} to match the corresponding curve ID from the luminance curve bank and retrieve the luminance feature $\mathbf{M}_{L}^{n}$ from the corresponding slot in $\mathbf{M}_{L}$. 
Based on the $\mathbf{M}_{L}^{n}$, we guide the model to extract complementary structural features from the depth map $\mathbf{I}_{dep}$. 
We first encode the $\mathbf{I}_{dep}$: $\mathbf{F}_{en}^{de} = \mathcal{E}\left(\mathbf{I}_{dep}\right)$.

Then, we apply a linear layer to $\mathbf{M}_{L}^{n}$ to compute ${Q_{lu}} $, and apply another linear layer to $\mathbf{F}_{en}^{de}$ to compute ${K_{de}}$ and ${V_{de}}$.
Using ${Q_{lu}}$ to query ${K_{de}}$ yields the matching scores between the luminance feature $\mathbf{M}_{L}^{n}$ and the structural features $\mathbf{F}_{en}^{de}$:
\begin{equation}
\begin{array}{l}
Q\cdot K = \underbrace{{Q_{lu}} \in \mathbb{R}^{HW \times C}}_{\text{luminance}} \times \underbrace{{K_{de}} \in \mathbb{R}^{HW \times C}}_{\text{structure}}, \hfill \\[16pt]
Score\in \mathbb{R}^{HW \times C} = \text{Softmax}\left(Q\cdot K\right).\\
\end{array}
\label{eqa}
\end{equation}

These scores highlight the varying attention that the luminance feature $\mathbf{M}_{L}^{n}$ assigns to valid and invalid information within the depth map features.
Then, ${V_{de}}$ is modulated according to these scores and reshaped to obtain the structural compensation feature $\mathbf{F}_{en}^{s}$:
\begin{equation}
\begin{array}{l}
\mathbf{F}_{en}^{s}\in \mathbb{R}^{H\times W \times C} = \text{Reshape}\left(Score \times {V_{de}} \right).\\
\end{array}
\label{qkv}
\end{equation}

To constrain the above structural modeling process, we introduce a Canny map. For the input image set $\{\mathbf{I}_{1},\mathbf{I}_{2},...,\mathbf{I}_{N}\}$, we first use a histogram to identify the brightest image $\mathbf{I}_{max}$, and process it with a Canny detector to generate a $\mathbf{S}_{can}$, which serves as a reference for the structure modeling. 
Under the constraint of the $\mathbf{S}_{can}$, $\mathbf{F}_{en}^{s}$ is decoded using a decoder to produce the structural map:
\begin{equation}
\begin{array}{l}
\mathbf{S}_{can}^{dep} = \mathcal{D}\left(\mathbf{F}_{en}^{s}\right)\leftarrow\mathcal{L}_{bce}\left(\mathbf{S}_{can}^{dep},\mathbf{S}_{can}\right), \hfill \\
\end{array}
\end{equation}
where $\mathcal{L}_{bce}$ is the binary cross entropy. $\mathbf{S}_{can}$ is the structure map extracted from $\mathbf{I}_{max}$ using the Canny detector.
\begin{figure*}[t]
  \centering
  \includegraphics[height=5.4cm]{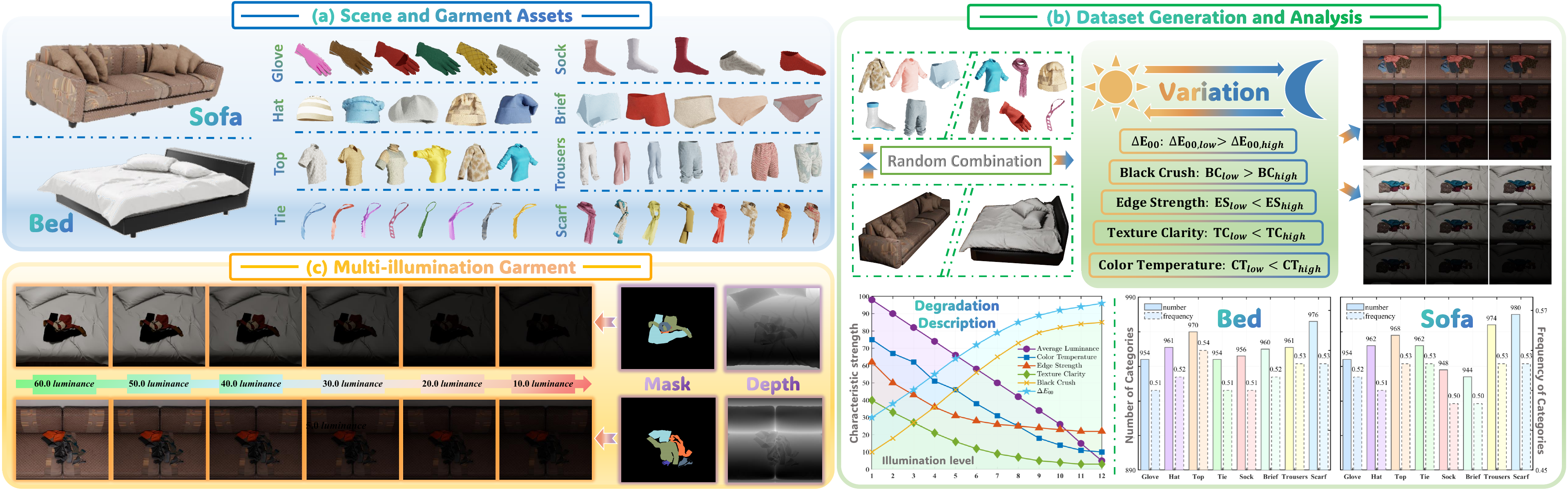}
  \vspace{-0.3cm}
  \caption{
  Construction of our MIGG dataset. (a) shows the household scenes and eight categories of garment assets.
  (b) shows that we obtain images under different illumination conditions by randomly combining the assets with the scenes. 
  We undergo degradation in multiple dimensions such as color difference ($\Delta{\mathbf{E}_{00}}$), black crush, texture clarity, and color temperature, rather than simple luminance attenuation. 
  The lower part of (b) describes the dataset’s degradation status and the number/frequency of each garment category.
  (c) presents the captured garment images, depth map and semantic mask map. 
  ``\textbf{Luminance}" is calculated by averaging the histograms.
  }
  \vspace{-0.6cm}
  \label{data}
\end{figure*}

Similar to Eq.~\ref{eq4}, we introduce a Structural Response Library (SRL: $\mathbf{M}_{S}=\{\mathbf{M}^{1}_{S},\mathbf{M}^{2}_{S},...,\mathbf{M}^{N}_{S}\}$) to store the structural compensation features $\mathbf{F}_{en}^{s}$ in the corresponding slot of $\mathbf{M}_{S}$ according to the current curve ID, and update them using the EMA strategy.
$\mathbf{M}_S$ will support the structural modeling in the grasp point prediction described in Sec.~\ref{s33}.
\subsection{Semantic Mask and Grasp Point Prediction}
\label{s33}
We divide grasp point prediction into two closely connected stages: identifying the semantic mask regions of garments (Sec.~\ref{331}) and determining the optimal grasp points based on these semantic mask regions (Sec.~\ref{332}).
\subsubsection{Garment Category Recognition}
\label{331}
Garment grasping involves diverse garment types, while existing method~\cite{R10} primarily relies on point cloud signals to model grasp points, neglecting semantic category information and thus struggling with multi-category grasping. 
In contrast, our method uses semantic information to enable garment grasping tailored to different categories.
For an input captured under arbitrary illumination, we obtain the corresponding curve ID from the curve library in the same manner as Eq.~\ref{eq2}.
Using the curve ID, we retrieve the associated luminance feature $\mathbf{M}_L$ and structural feature $\mathbf{M}_S$ from the luminance and structural response libraries.
To enable complementary enhancement of luminance and structure based on these retrieved features, we perform luminance–structure feature decomposition on the input $\mathbf{I}_{n}$:
\begin{equation}
\begin{array}{l}
\mathbf{I}_{L}\in\mathbb{R}^{H\times W\times3},\mathbf{I}_{S}\in\mathbb{R}^{H\times W\times1} = \mathcal{N}_{Retinex}\left( \mathbf{I}_{n}\right), \hfill \\
\end{array}
\end{equation}
where $\mathbf{I}_{L}$ and $\mathbf{I}_{S}$ denote the decomposed luminance and structural maps, respectively.
We use the network from ~\cite{R50} as the image decomposition network $\mathcal{N}_{Retinex}\left(\cdot\right)$.

Then, we use the encoder to encode $\mathbf{I}_{L}$ and $\mathbf{I}_{S}$ to obtain the luminance feature $\mathbf{F}_{L}$ and the structural feature $\mathbf{F}_{S}$.
Based on the feature $\mathbf{F}_{L}$, we apply a linear layer to compute $\mathbf{F}_{L}$ as $Q_{lu}$, and compute $K_{m}$ and $V_{m}$ from the feature $\mathbf{M}_L$.
The same with Eq.~\ref{eqa}, we use ${Q}_{lu}$ to query ${K}_m$, we obtain the matching scores between the feature $\mathbf{F}_{L}$ and the features $\mathbf{M}_{L}$ from the response library.
These scores reflect the missing luminance features in the input, which is then modulated by multiplying with ${V}_m$ to generate the enhanced luminance feature $\mathbf{F}_{L}^{en}$.
Similarly, we compute the matching scores between the structural feature $\mathbf{F}_{S}$ and the corresponding features $\mathbf{M}_{S}$ in the structural response library, thereby obtaining the enhanced structural feature $\mathbf{F}_{S}^{en}$.

Finally, we concatenate the enhanced structural and luminance features, followed by feature mapping and decoding operations to obtain the semantic mask map $\mathcal{M}_{m}$:
\begin{equation}
\begin{array}{l}
\mathcal{F} = \text{MLP}\left(\text{Concatenate}\left(\mathbf{F}_{S}^{en},\mathbf{F}_{L}^{en}\right)\right), \hfill\\[4pt]
\mathcal{M}_{m}=\mathcal{D}\left(\text{Reshape}\left(\mathcal{F}\right)\right)\leftarrow \mathcal{L}_{ce}\left({\mathcal{M}_{m}},gt\right),
\end{array}
\end{equation}
where $\mathcal{L}_{ce}$ is the cross-entropy loss used to constrain the decoder in generating the semantic mask, and \textit{gt} is the label.
\subsubsection{Category-Specific Grasp Point Generation} 
\label{332}
\begin{table*}[t]
  \centering
  \vspace{-0.1cm}
 \scalebox{0.695}{
    \begin{tabular}{p{3.4em}||p{2.6em}<{\centering}p{2.4em}<{\centering}p{2.5em}<{\centering}p{4em}<{\centering}||p{2.6em}<{\centering}p{2.4em}<{\centering}p{2.5em}<{\centering}p{4em}<{\centering}||p{2.6em}<{\centering}p{2.4em}<{\centering}p{2.5em}<{\centering}p{4em}<{\centering}||p{2.6em}<{\centering}p{2.4em}<{\centering}p{2.5em}<{\centering}p{4em}<{\centering}}
    \toprule
    \toprule
   \rowcolor[rgb]{ .8,  .8,  .8}\multicolumn{1}{l||}{\textbf{Class}} &\multicolumn{4}{c||}{\textbf{Luminance: 90 -- 120}} & \multicolumn{4}{c||}{\textbf{\textbf{Luminance: 60 -- 90}}} & \multicolumn{4}{c||}{\textbf{\textbf{Luminance: 30 -- 60}}} & \multicolumn{4}{c}{\textbf{\textbf{Luminance: 0 -- 30}}}  \\
    \midrule
        \rule{0pt}{14pt}\raisebox{0.8ex}{
      ------ } & \raisebox{0.8ex}{SegMiF} & \raisebox{0.8ex}{MRFS} & \raisebox{0.8ex}{AMDA} & \raisebox{0.8ex}{\textbf{Ours}} &
      \raisebox{0.8ex}{SegMiF} & \raisebox{0.8ex}{MRFS} & \raisebox{0.8ex}{AMDA} & \raisebox{0.8ex}{\textbf{Ours}}  &
      \raisebox{0.8ex}{SegMiF} & \raisebox{0.8ex}{MRFS} & \raisebox{0.8ex}{AMDA} & \raisebox{0.8ex}{\textbf{Ours}}  &
      \raisebox{0.8ex}{SegMiF} & \raisebox{0.8ex}{MRFS} & \raisebox{0.8ex}{AMDA} & \raisebox{0.8ex}{\textbf{Ours}} 
       \\
     \cdashline{1-17}
\rowcolor[rgb]{ .9,  .9,  .9}\textbf{Glove}& 74.9\% & 77.9\% & 76.6\% & \textbf{84.4\%} & 70.1\% & 71.1\% & 75.5\% & \textbf{84.2\%} & 61.9\% & 62.5\% & 70.7\% & \textbf{83.1\%} & 62.3\% & 61.8\% & 68.5\% & \textbf{81.7\%} \\
\textbf{Hat}& 82.8\% & 80.1\% & 81.5\% & \textbf{86.4\%} & 74.6\% & 78.7\% & 80.3\% & \textbf{85.8\%} & 71.8\% & 70.6\% & 75.9\% & \textbf{85.5\%} & 65.4\% & 68.2\% & 71.3\% & \textbf{84.7\%} \\
\rowcolor[rgb]{ .9,  .9,  .9}\textbf{Scarf}& 76.0\% & 82.2\% & 79.1\% & \textbf{85.7\%} & 73.3\% & 75.1\% & 77.0\% & \textbf{86.3\%} & 65.4\% & 65.7\% & 73.1\% & \textbf{85.7\%} & 66.3\% & 63.5\% & 68.6\% & \textbf{85.6\%} \\
\textbf{Sock}& 70.9\% & 67.2\% & 72.8\% & \textbf{83.5\%} & 66.2\% & 65.7\% & 68.0\% & \textbf{82.2\%} & 62.3\% & 58.9\% & 65.5\% & \textbf{80.3\%} & 56.9\% & 61.8\% & 59.3\% & \textbf{79.2\%} \\
\rowcolor[rgb]{ .9,  .9,  .9}\textbf{Tie} & 72.5\% & 80.3\% & 79.9\% & \textbf{84.7\%} & 70.7\% & 73.9\% & 74.4\% & \textbf{84.1\%} & 64.1\% & 66.2\% & 71.1\% & \textbf{82.1\%} & 60.8\% & 65.3\% & 66.9\% & \textbf{81.4\%} \\
\textbf{Top}& 73.5\% & 70.1\% & 75.9\% & \textbf{86.2\%} & 71.6\% & 73.2\% & 72.8\% & \textbf{85.8\%} & 74.4\% & 73.4\% & 72.4\% & \textbf{85.2\%} & 71.4\% & 70.4\% & 72.5\% & \textbf{84.4\%}\\
\rowcolor[rgb]{ .9,  .9,  .9}\textbf{Trousers}& 68.8\% & 73.9\% & 74.1\% & \textbf{84.3\%} & 65.4\% & 69.6\% & 70.4\% & \textbf{82.9\%} & 61.1\% & 63.9\% & 67.5\% & \textbf{82.7\%} & 60.4\% & 61.7\% & 65.0\% & \textbf{81.5\%} \\
\textbf{Brief}& 78.2\% & 78.4\% & 79.1\% & \textbf{83.5\%} & 72.2\% & 73.0\% & 75.9\% & \textbf{83.3\%} & 67.2\% & 67.7\% & 76.5\% & \textbf{82.1\%} & 57.1\% & 62.5\% & 65.7\% & \textbf{80.2\%} \\
\midrule
\rowcolor[rgb]{ .8,  .93,  1} \textbf{mIoU}& 74.7\% & 76.2\% & 77.3\% & \textbf{84.8\%\textcolor{red}{\scriptsize~+14\%}} & 70.5\% & 72.6\% & 74.4\% & \textbf{84.3\%\textcolor{red}{\scriptsize~+14\%}} & 66.0\% & 66.3\% & 71.5\% & \textbf{83.4\%\textcolor{red}{\scriptsize~+17\%}} & 62.5\% & 64.2\% & 67.3\% & \textbf{82.8\%\textcolor{red}{\scriptsize~+20\%}} \\
    \bottomrule
    \bottomrule
    \end{tabular}%
    }
    \vspace{-0.3cm}
    \caption{Quantitative comparison of semantic mask generation accuracy between our GraspALL and other baseline methods under different luminance levels. The best results are highlighted in \textbf{bold}, and performance improvements are shown in \textcolor{red}{\textbf{red}}.}
    \vspace{-0.2cm}
  \label{t:1}%
\end{table*}%
\begin{table*}[t]
  \centering
  \vspace{-0.1cm}
 \scalebox{0.711}{
    \begin{tabular}{p{3.5em}||p{2.9em}<{\centering}p{3em}<{\centering}p{2.5em}<{\centering}p{2.7em}<{\centering}p{4.1em}<{\centering}||p{2.9em}<{\centering}p{3em}<{\centering}p{2.5em}<{\centering}p{2.7em}<{\centering}p{4.1em}<{\centering}||p{2.9em}<{\centering}p{3em}<{\centering}p{2.5em}<{\centering}p{2.7em}<{\centering}p{4.1em}<{\centering}}
    \toprule
    \toprule
   \rowcolor[rgb]{ .8,  .8,  .8}\multicolumn{1}{l||}{\textbf{Class}} &\multicolumn{5}{c||}{\textbf{Luminance: 80 -- 120}} & \multicolumn{5}{c||}{\textbf{\textbf{Luminance: 40 -- 80}}} & \multicolumn{5}{c}{\textbf{\textbf{Luminance: 0 -- 40}}}  \\
    \midrule
    \rule{0pt}{14pt}
         \raisebox{0.8ex}{------}& \raisebox{0.8ex}{BiFCNet}& \raisebox{0.8ex}{SAM-M} & \raisebox{0.8ex}{ReKep} & \raisebox{0.8ex}{DarkSeg} & \raisebox{0.8ex}{\textbf{Ours}} &\raisebox{0.8ex}{BiFCNet}& \raisebox{0.8ex}{SAM-M} & \raisebox{0.8ex}{ReKep} & \raisebox{0.8ex}{DarkSeg} & \raisebox{0.8ex}{\textbf{Ours}} &\raisebox{0.8ex}{BiFCNet}& \raisebox{0.8ex}{SAM-M} & \raisebox{0.8ex}{ReKep} & \raisebox{0.8ex}{DarkSeg} & \raisebox{0.8ex}{\textbf{Ours}}\\
    \cdashline{1-16}
        \rowcolor[rgb]{ .9,  .9,  .9}\textbf{Glove} & 9/15 & 10/15 & 8/15 & 11/15 & \textbf{14/15} & 8/15 &  9/15  & 8/15  & 10/15 &\textbf{14/15} & 5/15  & 5/15 & 6/15 & 7/15 & \textbf{12/15} \\
          \textbf{Hat}& 11/15 & 10/15 & 9/15 & 13/15 & \textbf{15/15} & 9/15 &  8/15  & 8/15  & 9/15 & \textbf{13/15} & 6/15  & 7/15 & 6/15 & 8/15 & \textbf{13/15}  \\
          \rowcolor[rgb]{ .9,  .9,  .9}\textbf{Scarf}& 10/15 & 9/15 & 10/15 & 12/15 &
          \textbf{14/15} & 8/15 &  8/15  & 7/15  & 8/15 & \textbf{13/15} & 6/15  & 6/15 & 5/15 & 7/15 &\textbf{13/15}\\
      \textbf{Sock}& 9/15 & 7/15 & 10/15 & 11/15 & \textbf{13/15} & 7/15 &  7/15  & 7/15  & 9/15 & \textbf{13/15} & 5/15  & 5/15 & 6/15 & 7/15 & \textbf{11/15}  \\
              \rowcolor[rgb]{ .9,  .9,  .9}\textbf{Tie} & 9/15 & 10/15 & 8/15 & 10/15 & \textbf{13/15} & 6/15 &  5/15  & 6/15  & 7/15 & \textbf{11/15} & 3/15  & 5/15 & 5/15 & 6/15 & \textbf{10/15} \\
          \textbf{Top}&  10/15 & 8/15 & 11/15 & 13/15 &\textbf{15/15} & 8/15 &  9/15  & 9/15  & 11/15 & \textbf{14/15} & 8/15  & 8/15 & 9/15 & 10/15 & \textbf{14/15}  \\
          \rowcolor[rgb]{ .9,  .9,  .9}\textbf{Trousers}&  9/15 & 11/15 & 11/15 & 13/15 & \textbf{14/15} & 10/15 &  10/15  & 10/15  & 12/15 & \textbf{14/15} & 9/15  & 10/15 & 8/15 & 12/15 &  \textbf{14/15}  \\
      \textbf{Brief}& 7/15 & 9/15 & 9/15 & 11/15 & \textbf{14/15} & 7/15 &  6/15  & 8/15  & 10/15 & \textbf{14/15} & 6/15  & 6/15 & 6/15 & 7/15 &  \textbf{13/15}  \\
      \midrule
       \rowcolor[rgb]{ .8,  .93,  1} \textbf{mGSR}& 61.6\% & 59.2\% & 63.4\% & 78.3\% & \textbf{93.3\%\textcolor{red}{\scriptsize~+32\%}} & 52.4\% & 51.6\% & 52.4\% & 63.3\% & \textbf{88.3\%\textcolor{red}{\scriptsize~+36\%}} & 39.9\% & 43.3\% & 42.4\% & 53.3\% & \textbf{84.2\%\textcolor{red}{\scriptsize~+44\%}} \\
    \bottomrule
    \bottomrule
    \end{tabular}%
    }
    \vspace{-0.27cm}
    \caption{Comparison of garment grasping accuracy between our GraspALL and other baseline methods under different luminance levels.}
    \vspace{-0.6cm}
  \label{t:2}%
\end{table*}%
In Fig.~\ref{fig:f3}, based on the mask map $\mathcal{M}_m$, our goal is to identify the graspable regions of different garment categories and determine the optimal grasp point using the corresponding depth map.
Previous methods ~\cite{R7,R8,R9} typically define the grasp point as the center of each garment in $\mathcal{M}_m$, which reduces dragging during grasp execution. 
However, due to the high deformability of garments, the geometric center does not always correspond to graspable folds, often leading to unstable grasping.
To address this, as shown in Fig.~\ref{fig:f8}, we propose a depth-optimal search strategy that ensures both stability during the grasping process.

Given a mask $\mathcal{M}_m$ containing multiple garment classes \textit{C}, we first select the class $c^{*}$ with the largest area $\Omega_{c^{*}}$:
\begin{equation}
\begin{array}{l}
\Omega_{c^{*}}=\mathop {\text{argmax}}\limits_{c \in C}\left|\Omega_c \right|,
\end{array}
\end{equation}
where $\Omega_c$ denotes the pixel set of class $c$ in $\mathcal{M}_m$.
This ensures that grasping is performed on the dominant garment region. 
Within the selected region $\Omega_{c^{*}}$, we extract $k$ pixels with the smallest depth values (i.e., the closest to the camera) based on the corresponding area of the depth map, representing the most accessible surface points:
\begin{equation}
\begin{array}{l}
p_{1},p_{2},...,p_{k}= \text{Depth}_{top}\left(\Omega_{c^{*}}\right),
\end{array}
\end{equation}
where $p_{1},p_{2},...,p_{k}$ are the points where the depth value is optimal.
The above process effectively identifies geometrically salient points on the garment surface, such as wrinkles or protrusions, thereby enhancing grasping stability.

Following the~\cite{R7}, we compute the geometric center $p_{center}$ of the current semantic region $\Omega_{c^{*}}$ by fitting the minimum bounding rectangle. 
Then, among the candidate points $p_{1},p_{2},...,p_{k}$, we select the pixel closest to $p_{center}$ as the optimal grasping point $p_{o}$:
\begin{equation}
\begin{array}{l}
p_{o}= \mathop {\text{argmin}}\limits_{p \in p_{k}}||p-p_{center}||_2,
\end{array}
\end{equation}
where $||\cdot||_2$ is the euclidean distance.   
By searching for the grasping point across the entire semantic region, our method avoids dragging issues caused by off-center grasping while ensuring that the selected point lies in a structurally stable, wrinkled area.   
After completing this process, we iteratively repeat the above steps to grasp the remaining garment.
\section{Experiments}
\begin{figure*}[t]
  \centering
  \includegraphics[height=5.4cm]{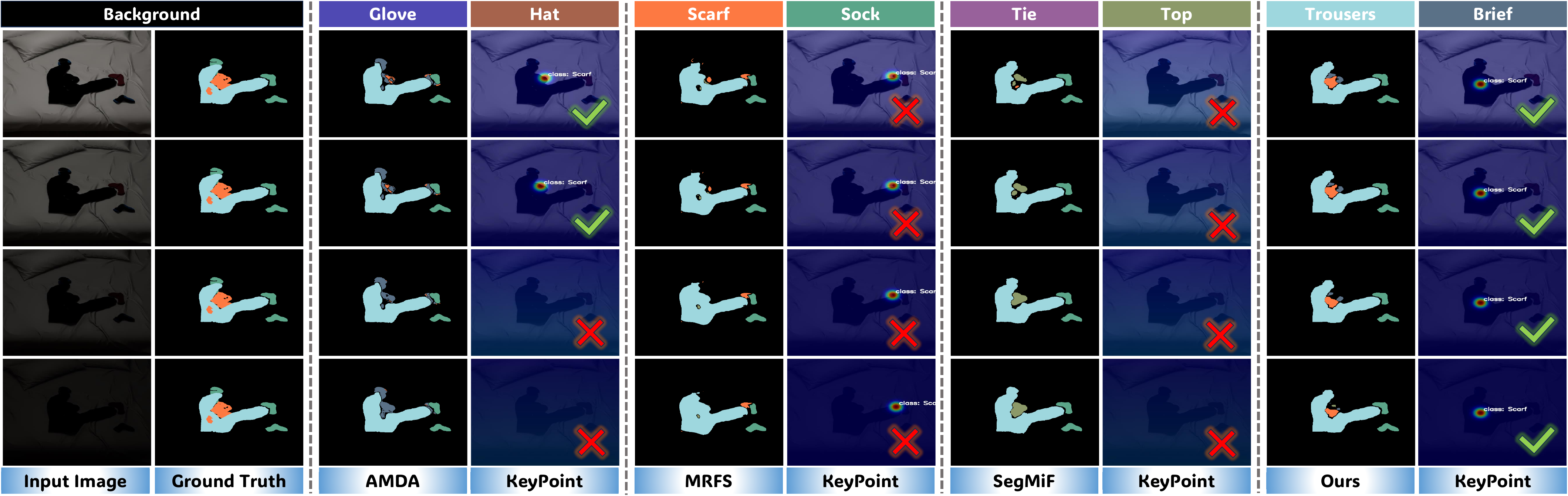}
  \vspace{-0.3cm}
  \caption{
   Comparison of semantic mask and grasping key points between our GraspALL and baselines under different luminance levels.
  }
  \vspace{-0.5cm}
  \label{fig:f5}
\end{figure*}
\subsection{Experimental Setup}
{\textbf{Dataset:}}
To overcome the scarcity of garment grasping datasets under varying illumination, as shown in Fig.~\ref{data}, we construct a \underline{\textbf{M}}ulti-\underline{\textbf{I}}llumination \underline{\textbf{G}}arment \underline{\textbf{G}}rasping (\textbf{MIGG}) dataset using NVIDIA Isaac Sim.
Unlike previous datasets ~\cite{R7,R40,R41} with limited garment types, uniform lighting, and simplified scenes, MIGG features controllable illumination and realistic household scenes.
We create two representative household scenes—a living-room sofa and a bedroom bed.
A garment asset library with eight categories (top, brief, glove, hat, tie, trousers, skirt, and sock) is developed, each modeled with realistic fabric properties and deformable dynamics.
Illumination is physically controlled through variations in intensity, direction, and color temperature, generating multiple brightness levels from normal to extreme low-light.
For each configuration, Isaac Sim captures synchronized RGB image, depth map, and semantic mask map at 512×512 resolution, yielding \textbf{15384} image triplets, split into \textbf{13008} for training and \textbf{2376} for testing.

\noindent{\textbf{Metrics:}}
We employ two complementary metrics to evaluate both semantic mask accuracy and mask-based garment grasping performance.
For semantic mask evaluation, \textbf{mIoU} (mean intersection over union)~\cite{R42} is used to measure accuracy.
For grasping evaluation, we design a multi-category garment grasping protocol, where each garment category is assigned a corresponding target basket. 
The model predicts grasp points from the semantic mask and executes grasp-and-place operations. 
A trial is considered successful if the garment is correctly grasped and placed into its target basket.
Each method is tested 15 times under every illumination level, and the \textbf{mGSR} (mean grasping success rate)~\cite{R43} is reported as the final evaluation metric.

\noindent{\textbf{Baselines:}}
GraspALL is trained using an NVIDIA 4090 GPU.
We compare GraspALL with representative baselines across both semantic mask generation and garment grasping.
For comparison of semantic mask, we select three multimodal fusion methods — SegMiF~\cite{R3}, MRFS~\cite{R2}, and AMDA~\cite{R1}.
For garment grasping comparison, we select BiFCNet~\cite{R4}, SAM-M~\cite{R5}, ReKep~\cite{R6}, and DarkSeg~\cite{R7} as baselines.
SegMiF, MRFS, AMDA, BiFCNet and DarkSeg will be retrained on our proposed dataset based on their original configuration.
SAM-M and ReKep use the official weights, and the prompt of ReKep is modified to: ``In low-light indoor environment, grasping the garment..''

\begin{itemize}
\item[$\bullet$]
BiFCNet is chosen to show the limitations of the traditional garment grasping method for illumination levels.
\end{itemize}

\begin{itemize}
\item[$\bullet$]
SAM-M and ReKep are selected to prove that even with the driving force of powerful large models, it remains difficult to handle complex low-light scenarios.
SAM-M is our modification of SAM based on ~\cite{R10}.
\end{itemize}
\begin{itemize}
\item[$\bullet$]
DarkSeg is chosen to verify that existing low-light garment grasping methods, though capable of addressing partial low-light conditions, struggle to handle diverse low-light situations under dynamic illumination.
\end{itemize}
\begin{table}[t]
  \centering
  \scalebox{0.7}{
    \begin{tabular}{p{4.1em}<{\centering}||p{2.4em}<{\centering}p{2.4em}<{\centering}p{2.4em}<{\centering}p{2.4em}<{\centering}p{2.4em}<{\centering}||p{3.1em}<{\centering}||p{3.1em}<{\centering}}
    \toprule
    \toprule
    \rowcolor[rgb]{ .8,  .8,  .8} \textbf{Model} & \textbf{PLC} & \textbf{LRL} & \textbf{SRL} & $\mathcal{L}_{sc}$ & $\mathcal{L}_{bce}$ & \textbf{mIoU} &\textbf{mGSR} \\
    \midrule
          \textbf{Model-1}& \ding{55}     &    &  && & 65.4\%  & 50.0\% \\
          \textbf{Model-2}&    & \ding{55}    & &&   & 71.3\%    & 72.5\% \\
          \textbf{Model-3}&    &     & \ding{55}   &&  &  68.5\% & 57.5\% \\
          \textbf{Model-4}&    &   &   & \ding{55}& & 64.9\%    & 50.2\% \\
          \textbf{Model-5}&    &   &   &  & \ding{55} & 68.7\%    & 70.4\% \\
    \midrule
      \rowcolor[rgb]{ .8,  .93,  1}  \textbf{Model-6} & {\ding{52}}& {\ding{52}}& {\ding{52}}   & \ding{52}  &\ding{52}  & \textbf{82.6\%}    & \textbf{88.3\%} \\
    \bottomrule
    \bottomrule
    \end{tabular}%
  }
    \vspace{-0.25cm}
    \caption{Ablation study of different components (Lum: 0-40).}
  \vspace{-0.7cm}
    \label{tab:3}
\end{table}%

\subsection{Quantitative Analysis}
Tab.~\ref{t:1} presents a quantitative comparison between our method and other multimodal models. Although other methods also leverage depth modality to enhance RGB features, they exhibit significant performance degradation under illumination variation. For instance, the MRFS suffers a 12\% drop in mIoU when transitioning from bright to low light. In contrast, GraspALL maintains stable performance across different brightness levels, with fluctuations under 2\%. This robustness stems from our proposed parametric luminance representation, which enables dynamic adaptation of feature distributions to illumination shifts, ensuring stable perception under diverse lighting conditions.

Tab.~\ref{t:2} reports the quantitative comparison of garment grasping performance. 
As shown, our GraspALL consistently outperforms all baselines across luminance ranges, achieving 83.3\% mGSR under extreme low-light conditions (0–40), representing a 31\% improvement over the second-best method. Even under medium brightness variations (40–120), GraspALL maintains a stable grasping success rate between 88.3\%–93.3\%, demonstrating strong robustness to illumination changes. These results confirm the effectiveness of our parametric luminance representation and adaptive compensation mechanism in achieving generalized and reliable grasping across complex lighting scenarios.
\begin{figure}[t]
  \centering
  \includegraphics[height=2.9cm]{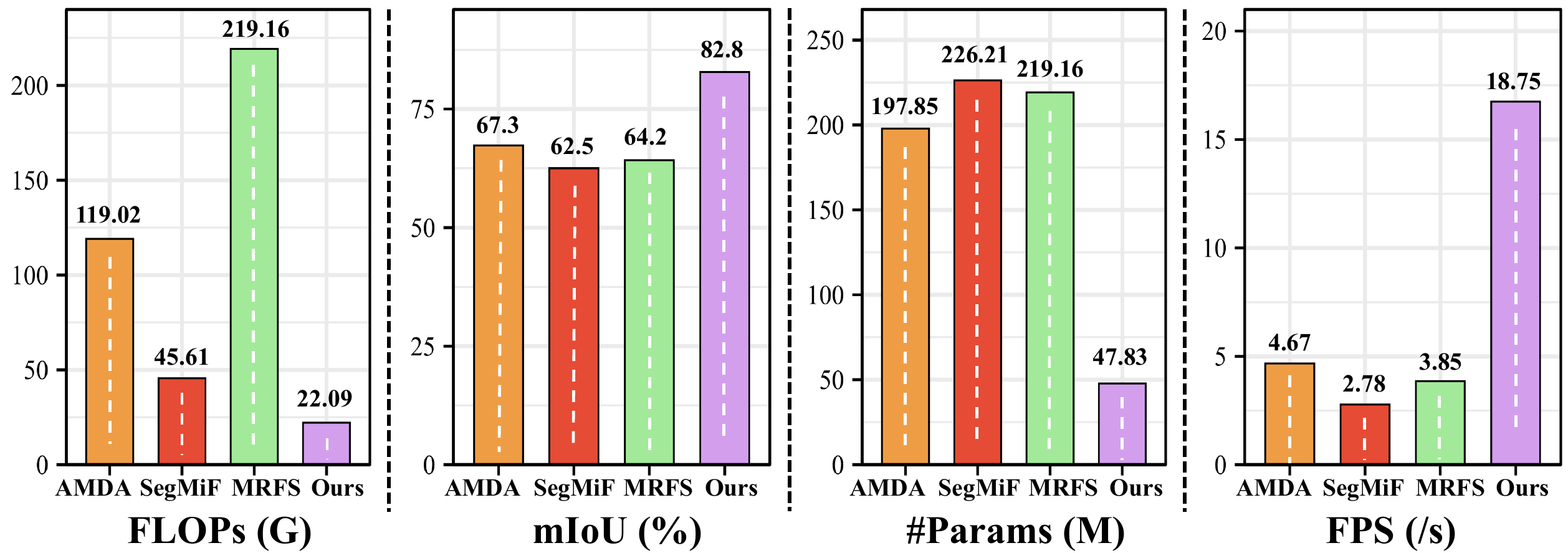}
  \vspace{-0.3cm}
  \caption{
  Comparisons of Parameters, FPS, mIoU and FLOPs. 
  }
  \vspace{-0.65cm}
  \label{fig:cm}
\end{figure}
\subsection{Qualitative Analysis}
Fig.~\ref{fig:f5} provides qualitative comparisons on semantic mask generation and grasp-point prediction. As illustrated, existing methods (e.g., SegMiF) produce blurred mask boundaries under varying illumination, failing to capture fine differences between garments and resulting in misplaced grasp points. 
In contrast, GraspALL preserves clear boundaries and consistent structural details even under low and dynamic lighting, with predicted grasp points concentrated on geometrically stable regions (e.g., wrinkles near the center). 
This demonstrates its superior precision and stability in grasping across challenging illumination conditions.
\subsection{Ablation Studies}
\label{abs}
We perform ablation studies on the Parametric Luminance Curve (PLC), Luminance Response Library (LRL), Structural Response Library (SRL), spectral consistency loss ($\mathcal{L}_{sc}$), and binary cross-entropy loss ($\mathcal{L}_{bce}$) of GraspALL.
The study involves six model variants:
\begin{figure}[t]
  \centering
  \includegraphics[height=3.5cm]{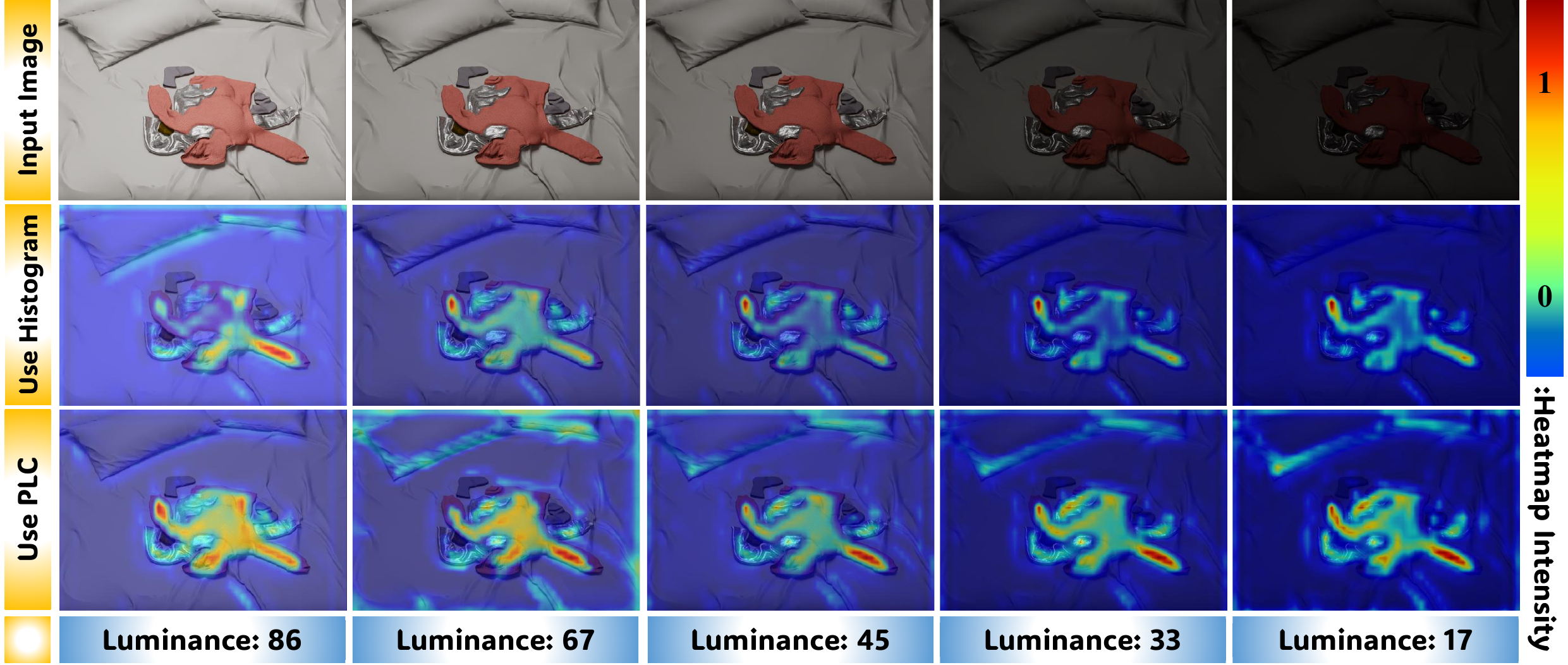}
  \vspace{-0.65cm}
  \caption{
  Differences in Grad-CAM at different luminance levels. 
  }
  \vspace{-0.66cm}
  \label{fig:ca}
\end{figure}
\textbf{Model-1}: PLC is removed, and the encoded features from both luminance and structural modeling are stored in fixed positions of LRL and SRL without luminance distinction.
\textbf{Model-2}: LRL is removed, and the structural modeling process stores the encoded features in SRL according to the luminance indicators produced by PLC.
\textbf{Model-3}: SRL is removed.
\textbf{Model-4}: $\mathcal{L}_{sc}$ is removed.
\textbf{Model-5}: $\mathcal{L}_{bce}$ is removed.
\textbf{Model-6}: Without any modification.
As shown in Tab.~\ref{tab:3}, all variants exhibit varying degrees of performance degradation compared to Model-6, confirming the effectiveness of each component. 
Model-1 shows the most significant performance drop, indicating that the absence of interpretable luminance estimation prevents explicit guidance for feature fusion between RGB and depth. 
The removal of SRL and LRL leads to performance decline, demonstrating that the complementary between structural and luminance features restores garment discriminability degraded by illumination changes.
Removing $\mathcal{L}_{sc}$ and $\mathcal{L}_{bce}$ results in performance drops. This is because the absence of $\mathcal{L}_{sc}$ eliminates the constraint signal for the PLC learning, while the lack of $\mathcal{L}_{bce}$ deprives the model of supervision for RGB-D fusion.
\subsection{Complexity Analysis}
We compare GraspALL with other models in terms of parameter, frames per second (FPS), and floating point operations per second (FLOPs). 
As shown in Fig.~\ref{fig:cm}, GraspALL not only achieves a higher mIoU but also delivers faster inference and a smaller model size than comparison methods. 
This is because we use the structural and luminance response libraries as intermediaries to decouple useful features from complex multimodal fusion, avoiding the intricate semantic alignment and feature interaction computations in multimodal fusion. 
When processing the input, the model only needs to estimate the illumination level via the PLC, then directly retrieve matching features from the two libraries for grasp point modeling—eliminating the need to repeatedly perform complex cross-modal fusion.
\subsection{Generalization Analysis of the PLC}
\label{46}
To evaluate the performance of PLC for unseen illumination, we employ Grad-CAM~\cite{R44} to visualize the encoder intermediate features of GraspALL.
For comparison, we replace the PLC with a traditional histogram mean method to generate luminance indices for input images.
As shown in Fig.~\ref{fig:ca}, GraspALL maintains stable attention across different illumination levels and even exhibits stronger structural focus under low light.
In contrast, GraspALL with the histogram method fails to produce more generalized interpretations for diverse inputs due to its non-learnable nature, thus showing significant attention divergence.
These results demonstrate that PLC can capture representative luminance patterns from input lighting and form a unified representation for arbitrary illumination, enabling accurate luminance interpretation even under unseen lighting conditions.
\begin{figure}[t]
  \centering
  \includegraphics[height=2.95cm]{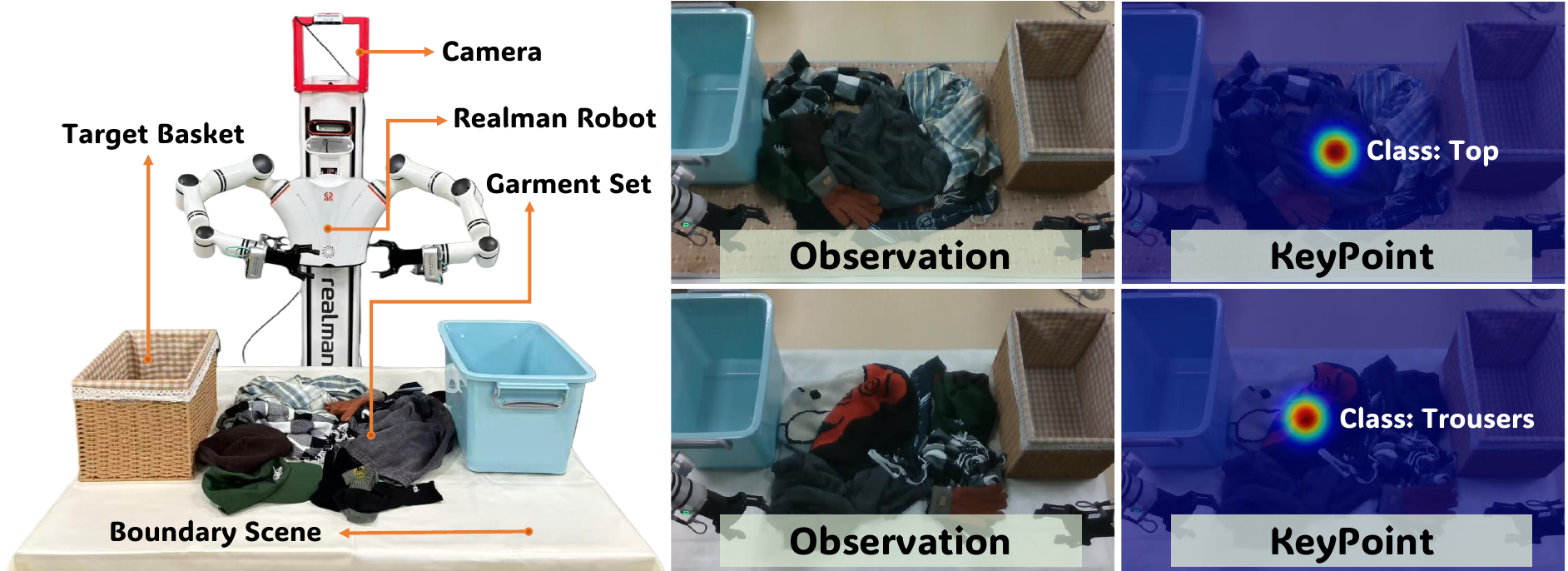}
  \vspace{-0.25cm}
  \caption{
   Setup and observation of real-world experiment.
  }
  \vspace{-0.35cm}
  \label{fig:rw}
\end{figure}
\begin{table}[t]
  \centering
  \scalebox{0.75}{
    \begin{tabular}{p{5em}<{\centering}||p{4em}<{\centering}|p{4em}<{\centering}|p{3em}<{\centering}|p{4em}<{\centering}|p{3em}<{\centering}}
    \toprule
    \toprule
    \rowcolor[rgb]{ .8,  .8,  .8}\textbf{Luminance}&\textbf{BiFCNet } & \textbf{SAM-M} & \textbf{ReKep} & \textbf{DarkSeg} & \textbf{Ours} \\
    \midrule
    Lu: 0 - 20 & 5 / 15 & 4 / 15 & 6 / 15  & 7 / 15 & \textbf{12 / 15} \\
   \rowcolor[rgb]{ .9,  .9,  .9}  Lu: 20 - 40  & 7 / 15 & 5 / 15 & 7 / 15  & 10 / 15 & \textbf{13 / 15} \\
    Lu: 40 - 60 & 7 / 15 & 7 / 15 & 9 / 15  & 11 / 15 & \textbf{13 / 15}\\ 
    \rowcolor[rgb]{ .9,  .9,  .9} Lu: 60 - 80 & 9 / 15  & 8 / 15 & 10 / 15 & 11 / 15  & \textbf{14 / 15}  \\
    \bottomrule
    \bottomrule
    \end{tabular}%
  }
    \vspace{-0.25cm}
    \caption{Real-world success rate under different luminance levels.}
  \vspace{-0.7cm}
    \label{tab:rw}%
\end{table}%
\subsection{Real-World Experiment}
To evaluate the real-world performance of GraspALL, we build a dataset of 1013 multi-illumination real-world garment images.
We follow the adaptation strategy ~\cite{R45}, transferring the weights of all models except ReKep and SAM-M, which were trained on the MIGG simulation dataset, to the real-world dataset, and then deploying them on the Realman robot for evaluation.
A RGB-Depth sensor, mounted above Realman's head and facing downward, is used for scene capture. 
Fig.~\ref{fig:rw} illustrates the real-world observation process, and Tab.~\ref{tab:rw} reports the superior grasping success rates achieved by GraspALL. Additional implementation details are provided in the supplementary material.

\section{Conclusion and Limitation}
For garment grasping under varying illumination, we propose GraspALL, an adaptive framework that leverages luminance curve and dual response libraries to achieve illumination-structure feature compensation. 
By encoding continuous illumination variations into quantifiable references and using these references to guide the feature complementarity between RGB and non-RGB modalities.
Experiments in both simulations and real-world deployments demonstrate that GraspALL improves grasping accuracy by 32-44\% under diverse illumination conditions.
Although our MIGG dataset is simulation-based, it is intentionally designed as a controlled benchmark that allows systematic analysis of illumination variations under physically consistent conditions. 
To further improve generalization, we plan to expand the real-world data in future work, incorporating more diverse household scenes and garment materials.
\section{Acknowledgements}
This work was supported by the National Natural Science Foundation of China (No. W2421093) and the International Cooperation Project of Jilin Province (No. 20250205079GH). This work was also supported by the National Science and Technology Council, Taiwan under Grant 114-2221-E-006-114-MY3. This work was supported in part by the National Natural Science Foundation of China (No. 62476110, No. U2341229), and the MSIT (Ministry of Science and ICT), Korea, under the ITRC (Information Technology Research Center) support program (IITP-2025-RS-2024-00437102).

{
    \small
    \bibliographystyle{ieeenat_fullname}
    \bibliography{main}
}

\clearpage
\setcounter{page}{1}
\maketitlesupplementary

This supplementary material presents additional experimental results that are omitted from the main paper due to the space limit. In this supplementary material, we provide:
\begin{itemize}

\item [$-$]
\underline{\textbf{1.}} Explanation of using RGB in extremely low-light conditions (in Sec.~\ref{s10}); 

\item[$-$]
\underline{\textbf{2.}} Explanation of the PLC (parametric luminance curve) generation process (in Sec.~\ref{s1});

\item[$-$]
\underline{\textbf{3.}} Analysis of the parameters $N$ and $R$ in PLC (in Sec.~\ref{s2});

\item[$-$]
\underline{\textbf{4.}} Real-world experiment deployment details and real-world images display (in Sec.~\ref{s3});

\item [$-$]
\underline{\textbf{5.}} Our solution strategy for dealing with low-quality depth maps (in Sec.~\ref{s4}); 

\item[$-$]
\underline{\textbf{6.}} Performance analysis for the proposed grasping strategy (i.e., depth-optimal search strategy) (in Sec.~\ref{s6}); 

\item[$-$]
\underline{\textbf{7.}} Analysis for the parameter $\alpha$ of the EMA (in Sec.~\ref{s7});

\item [$-$]
\underline{\textbf{8.}} Validation of GraspALL's generalization for other deformable objects (in Sec.~\ref{s8}); 

\item [$-$]
\underline{\textbf{9.}} Statistical analysis of grasping performance (in Sec.~\ref{s9}); 

\item [$-$]
\underline{\textbf{10.}} Robustness analysis of our GraspALL (in Sec.~\ref{s11}).
\end{itemize}

\section{Explanation of using RGB in extremely low-light condition}
\label{s10}
\begin{figure}[t]
  \centering
  \includegraphics[height=3.3cm]{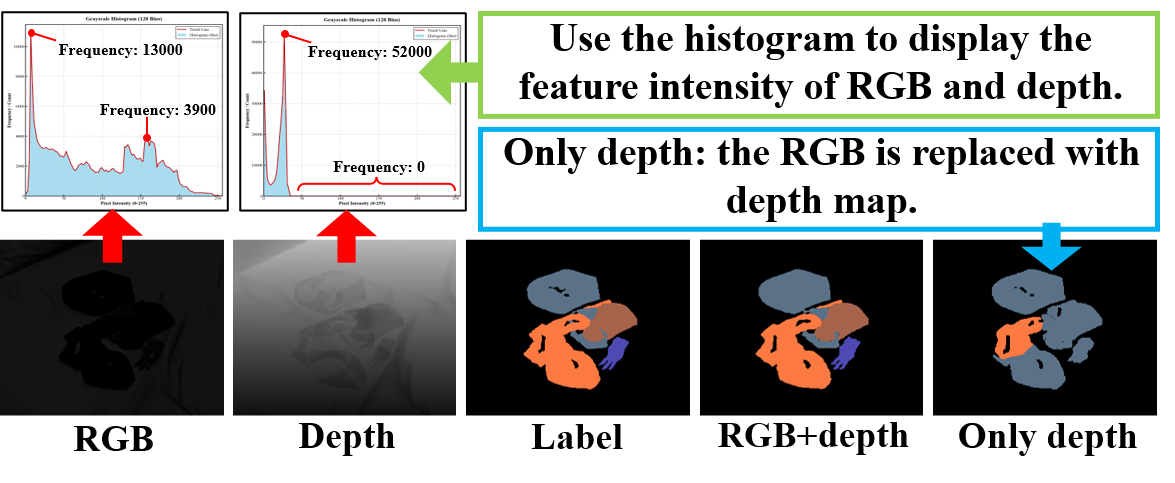}
  \vspace{-0.4cm}
  \caption{
Explanation of using RGB in low-light conditions.
  }
  \vspace{-0.3cm}
  \label{fig:10}
\end{figure}

In extremely low-light conditions, our GraspALL still uses RGB images to provide faint but crucial visual information, rather than relying solely on depth map information.
Depth mainly captures geometry but lacks discriminative cues such as color, material (Fig.~\ref{fig:10} upper left). Since garments are highly deformable, different categories exhibit similar shapes, so using depth alone leads to class confusion. 
In contrast, low-light RGB, though weak, still retains useful semantic cues that complement depth feature by semantically consistent fusion. 
In Fig.~\ref{fig:10}, training with depth alone leads to severe misclassification.

\section{Explanation of the PLC Generation Process}
\label{s1}
In GraspALL, we propose a parametric luminance curve (PLC), which learns a set of curves capable of representing arbitrary illumination from multiple luminance inputs, thereby accurately estimating the luminance of the input. 

The PLC forces curve IDs to indicate more accurate positions by calculating the feature consistency between the features generated by the encoder and those in the response library guided by curve IDs. This further guides the model to adjust curve parameters to generate more accurate IDs. However, considering that curve ID generation is discrete and non-differentiable, to enable end-to-end differentiable learning for the PLC and allow gradients to flow to PLC parameters, we adopt a differentiable processing mechanism in practice. 
First, we calculate the negative distance for each curve and perform temperature-scaled softmax:
\begin{equation}
\begin{array}{l}
\displaystyle
w_n = \frac{{\exp\left( -||\mathbf{H}-\mathbf{C}\left(\mathbf{P_{n}}\right)||/\tau \right)}} {{\sum _{m} {\exp\left( -||\mathbf{H}-\mathbf{C}\left(\mathbf{P_{m}}\right)||/\tau \right) } }},\\
\end{array}
\label{ee4}
\end{equation}
where \(\tau\) is the temperature coefficient, which controls the smoothness of the soft-hard distribution.

Then, we use the softmax weight $w_n$ to perform a weighted sum of the corresponding features in the luminance response library $\textbf{M}_L$:
\begin{equation}
\begin{array}{l}
\mathbf{M}_L = \sum\limits_{n = 1}^N w_n\mathbf{M}^{n}_{L}.\\
\end{array}
\label{ee4}
\end{equation}

Finally, we calculate the spectral consistency loss (\(\mathcal{L}_{sc}\)):
\begin{equation}
\begin{array}{l}
\displaystyle
\mathcal{L}_{sc} = ||\mathbf{F}^{n}_{en} - \mathbf{M}^{n}_{L}||_1,
\\
\end{array}
\label{ee4}
\end{equation}
where $||\cdot||_1$ represents the L1 loss.
Through the above differentiable processing, gradients can be backpropagated to \(\mathbf{F}_n^{en}\) and parameters \(\mathbf{P}_n\) in the PLC via $\mathbf{M}_L$.
\begin{table}[t]
  \centering
  \vspace{0.1cm}
  \scalebox{0.8}{
    \begin{tabular}{p{7.5em}|p{6.4em}<{\centering}|p{6.4em}<{\centering}|p{3.9em}<{\centering}}
    \toprule
    \toprule
 \rowcolor[rgb]{ .8,  .8,  .8} \textbf{Method}& \textbf{Luminance$<$20}   & \textbf{Luminance$<$40}  & \textbf{\#Params} \\
    \midrule
    $N=6,$ \ \ $R=256$ & mIoU:80.8\% & mIoU:81.5\%   & 34.53 \\
    \rowcolor[rgb]{ .8,  .93,  1} $N=12,R=256$ & mIoU:\textbf{82.2\%}  & mIoU:\textbf{83.1\%}  & \textbf{47.83} \\
    $N=18,R=256$ & mIoU:82.5\%  & mIoU:83.7\%  & 61.13 \\
    \midrule
    $R=128,N=12$ & mIoU:80.3\% & mIoU:80.9\%  & 47.83 \\
    \rowcolor[rgb]{ .8,  .93,  1} $R=256,N=12$ & mIoU:\textbf{82.2\%}  & mIoU:\textbf{83.1\%}  & \textbf{47.83} \\
    $R=512,N=12$ & mIoU:81.4\%  & mIoU:82.0\% & 47.83 \\
    \bottomrule
    \bottomrule
    \end{tabular}%
    }
    \vspace{-0.2cm}
  \caption{The effect of different $N$ and $R$ on model performance.}
  \vspace{-0.6cm}
  \label{tab:1}%
\end{table}%

Fig.~\ref{fig:1} visually presents the set of curves generated by the parametric luminance curve (PLC) after training.
As can be seen from Fig.~\ref{fig:1}, different curves have distinct focuses in representing luminance, enabling a comprehensive evaluation of the input luminance intensity.
\section{Analysis of the Parameters $N$ and $R$ in PLC}
\label{s2}
The PLC setup includes two key parameters: one is the number of luminance curves, denoted as $N$, and the other is the number of nodes per curve, denoted as $R$. 
For $N$, we set \(N = 12\), considering the symmetric illumination changes from dark to bright and then bright to dark over 24 hours a day. 
For $R$, we set \(R = 256\) to learn sufficient yet non-redundant luminance features from the input.

To verify the rationality of the settings for $N$ and $R$, we compared the impacts of different $N$ and $R$ values on model performance. 
The experimental results are presented in Tab.~\ref{tab:1}. 
As shown in Tab.~\ref{tab:1}, when \(N < 12\), the model accuracy degrades. 
This is because a smaller number of curves leads to vague illumination estimation by the PLC, failing to provide accurate guidance for the subsequent interactive enhancement of luminance and structural features. 
When \(N > 12\), although the model performance improves slightly, the increased number of curves requires more storage space to expand the luminance and structural response libraries, resulting in a significant increase in model complexity. 
Thus, \(N = 12\) achieves a better balance between accuracy and model complexity. Regarding the setting of $R$, Tab.~\ref{tab:1} indicates that model performance declines when \(R < 256\) or \(R > 256\). 
This is because an excessive $R$ causes the curves to overfit to a specific luminance pattern, while an insufficient $R$ makes it difficult for the curves to form a unified representation for diverse inputs. 
Therefore, \(R = 256\) enables the sufficient learning of representative patterns from different luminance inputs.
\begin{figure*}[t]
  \centering
  \includegraphics[height=10cm]{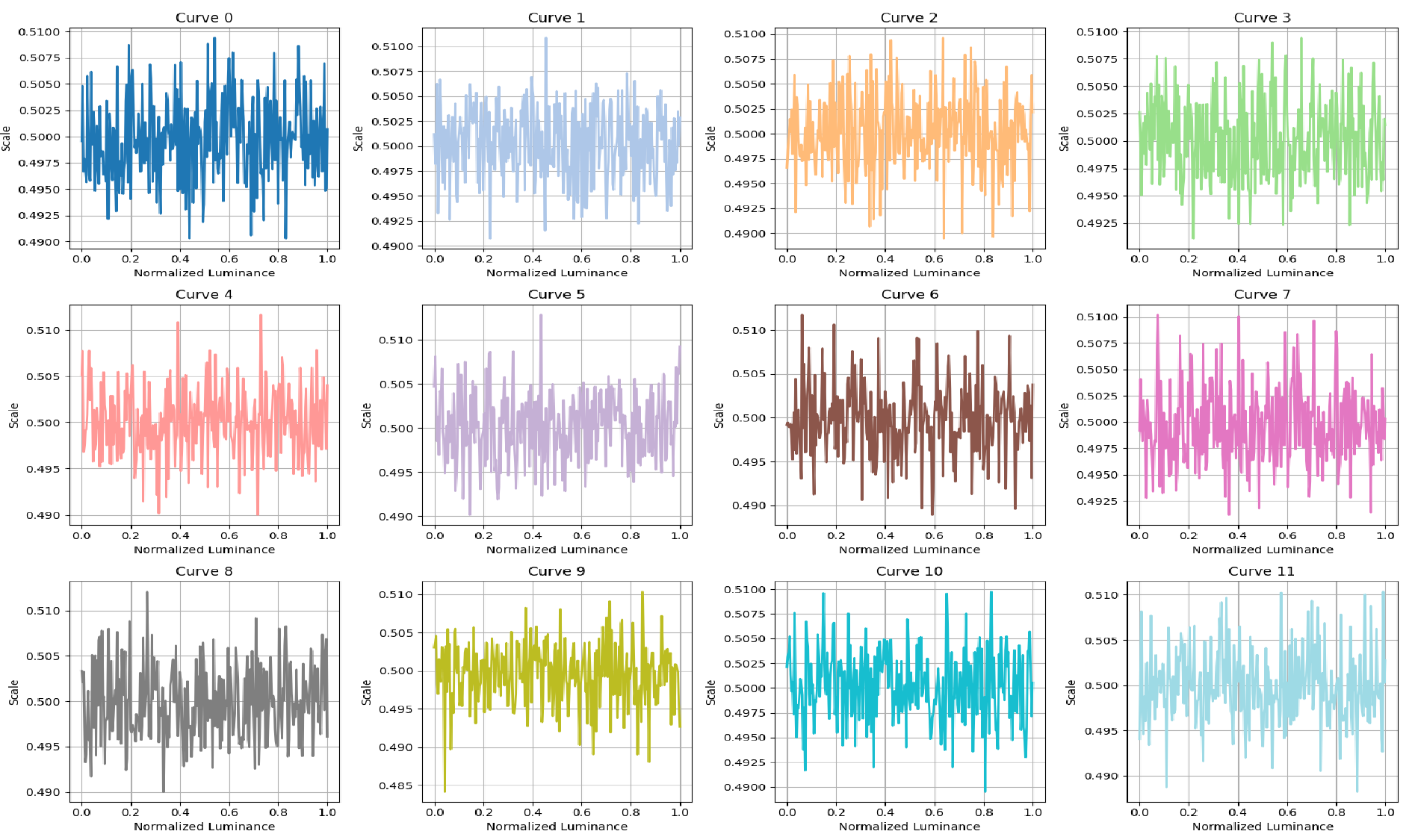}
  \vspace{-0.25cm}
  \caption{
Visualize the different luminance curves in the trained Parametric Luminance Curves.
  }
  \vspace{-0.35cm}
  \label{fig:1}
\end{figure*}
\section{Real-World Experiment Deployment}
\label{s3}
\begin{figure*}[t]
  \centering
  \includegraphics[height=8.4cm]{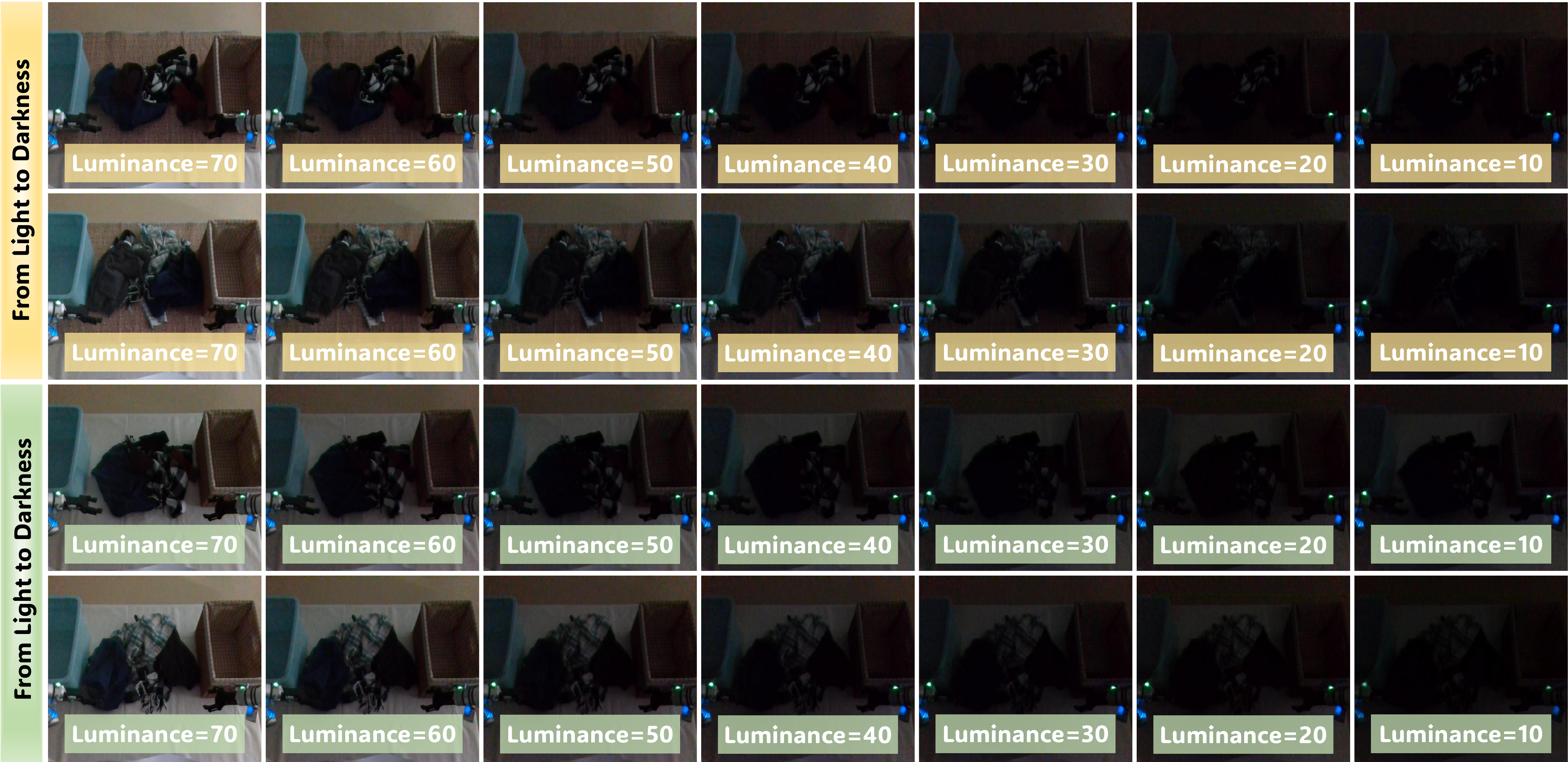}
  \vspace{-0.25cm}
  \caption{
Garment images under different illumination conditions collected in real-world scenarios.
  }
  \vspace{-0.5cm}
  \label{fig:2}
\end{figure*}

To verify the performance of GraspALL in real-world scenarios, we collected a dataset named RealData containing 1013 real-world captured images, as shown in Fig.~\ref{fig:2}. To enhance the sample diversity across different illumination levels, we systematically acquired garment images under varying illumination conditions (illumination range: 10–70) in a controlled environment by adjusting the curtain opening. The garment categories in RealData are consistent with those in the synthetic dataset MIGG.
\begin{table}[t]
  \centering
  \scalebox{0.7}{
    \begin{tabular}{p{4.5em}<{\centering}||p{4.9em}<{\centering}|p{5.5em}<{\centering}|p{4em}<{\centering}|p{4.7em}<{\centering}|p{2.8em}<{\centering}}
    \toprule
    \toprule
    \rowcolor[rgb]{ .8,  .8,  .8}\textbf{Luminance}&\textbf{BiFCNet~\cite{R4}} & \textbf{SAM-M~\cite{R5}} & \textbf{ReKep~\cite{R6}} & \textbf{DarkSeg~\cite{R7}} & \textbf{Ours} \\
    \midrule
    Lu: 0 - 20 & 5 / 15 & 4 / 15 & 6 / 15  & 7 / 15 & \textbf{12 / 15} \\
   \rowcolor[rgb]{ .9,  .9,  .9}  Lu: 20 - 40  & 7 / 15 & 5 / 15 & 7 / 15  & 10 / 15 & \textbf{13 / 15} \\
    Lu: 40 - 60 & 7 / 15 & 7 / 15 & 9 / 15  & 11 / 15 & \textbf{13 / 15}\\ 
    \rowcolor[rgb]{ .9,  .9,  .9} Lu: 60 - 80 & 9 / 15  & 8 / 15 & 10 / 15 & 11 / 15  & \textbf{14 / 15}  \\
    \bottomrule
    \bottomrule
    \end{tabular}%
  }
    \vspace{-0.2cm}
    \caption{Real-world grasping success rate of different methods.}
  \vspace{-0.5cm}
    \label{tab:2}%
\end{table}%
\begin{table}[t]
  \centering
  \vspace{0.18cm}
  \scalebox{0.7}{
    \begin{tabular}{p{6em}<{\centering}||p{5.7em}<{\centering}|p{5.7em}<{\centering}|p{5.7em}<{\centering}|p{3.5em}<{\centering}}
    \toprule
    \toprule
    \rowcolor[rgb]{ .8,  .8,  .8}\textbf{Luminance}&\textbf{SegMiF~\cite{R3}} & \textbf{MRFS~\cite{R2}} & \textbf{AMDA~\cite{R1}} & \textbf{Ours} \\
    \midrule
    Lu: 0 - 20 & 51.2\% & 52.9\% & 55.1\%  & \textbf{70.5\%} \\
   \rowcolor[rgb]{ .9,  .9,  .9}  Lu: 20 - 40  & 55.8\% & 58.3\% & 59.5\% & \textbf{71.3\%} \\
    Lu: 40 - 60 & 59.4\% & 62.7\% & 63.8\% & \textbf{72.4\%}\\ 
    \rowcolor[rgb]{ .9,  .9,  .9} Lu: 60 - 80  & 64.3\% & 65.6\% & 66.3\% & \textbf{74.3\%}  \\
    \bottomrule
    \bottomrule
    \end{tabular}%
  }
    \vspace{-0.2cm}
    \caption{Accuracy rate of mask generation by different methods.}
  \vspace{-0.25cm}
    \label{tab:3}%
\end{table}%

\begin{table}[t]
    \centering
     \scalebox{0.8}{
    \begin{tabular}{c|cc|cc}
        \toprule
        \toprule
        & \multicolumn{2}{c|}{\textbf{GraspALL (\textcolor{red}{mGSR})}} & \multicolumn{2}{c}{\textbf{DarkSeg (\textcolor{red}{mGSR})}} \\
        \midrule
        
        \rowcolor{gray!12}
        \textbf{Luminance} & \textbf{with ST} & \textbf{without ST} & \textbf{with ST} & \textbf{without ST} \\
        
        $0 - 20$ & 80.0\% & 66.7\% & 46.7\% & 33.3\% \\
        \midrule
        
        \rowcolor{gray!12}
        \textbf{Luminance} & \multicolumn{2}{c|}{\cellcolor{gray!12}\textbf{GraspALL (\textcolor{red}{mIoU})}} & \multicolumn{2}{c}{\cellcolor{gray!12}\textbf{DarkSeg (\textcolor{red}{mIoU})}} \\
        
        $0 - 20$ & 70.5\% & 61.4\% & 50.7\% & 36.9\% \\
        \bottomrule
         \bottomrule
    \end{tabular}}
    \vspace{-0.2cm}
    \caption{Analysis for our transfer strategy (sim-to-real).}
    \vspace{-0.5cm}
    \label{t222}
\end{table}

\begin{figure}[t]
  \centering
  \vspace{0.2cm}
  \includegraphics[height=3.45cm]{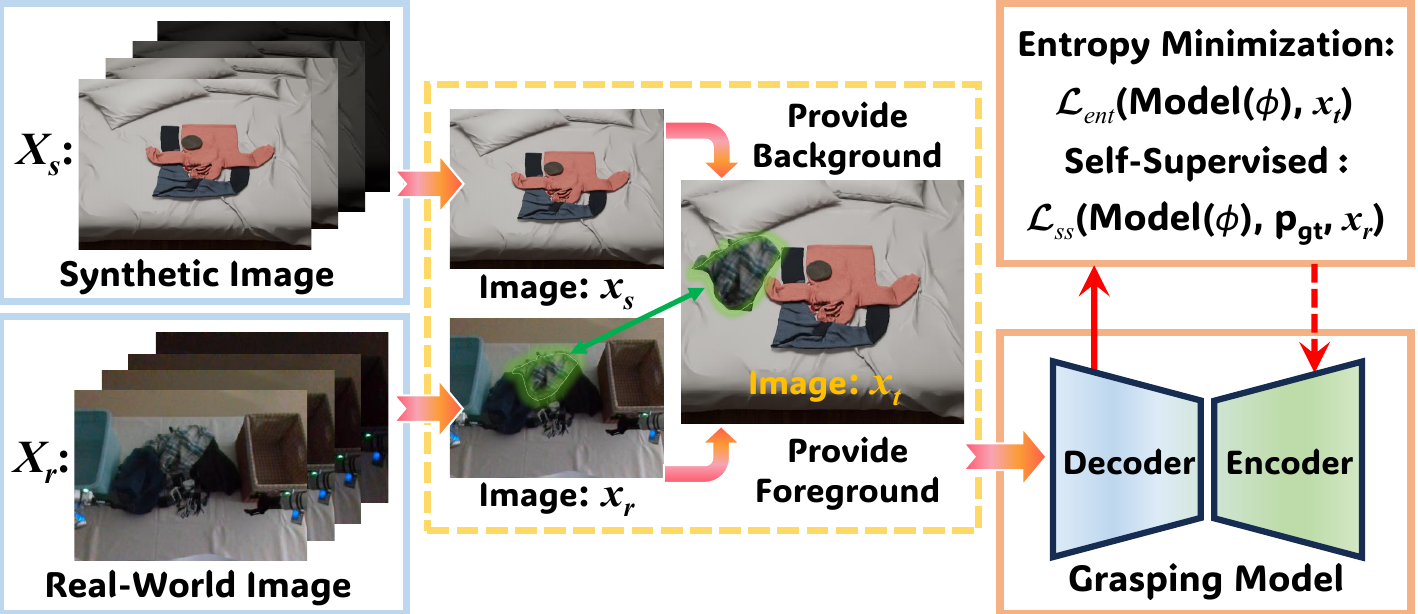}
  \vspace{-0.2cm}
  \caption{
We use the method in ~\cite{R45} to transfer the model trained on the synthetic dataset to the real-world dataset.
$\mathbf{Model}\left(\phi\right)$ denotes the model trained on synthetic data. $\mathbf{p}_{gt}$ represents the pseudo-labels generated by the model. The domain adaptation process is as follows: first, train the model using synthetic data; then, train the model using both synthetic and real data; finally, train the model using real data.
  }
  \vspace{-0.55cm}
  \label{fig:3}
\end{figure}

To achieve domain adaptation from the synthetic dataset MIGG to the real-image dataset RealData, we adopted the Fourier Domain Adaptation method proposed by Yang et al~\cite{R45}. As illustrated in Fig.~\ref{fig:3}, we performed Fast Fourier Transform (FFT) on synthetic and real images respectively, replaced the corresponding regions of the source domain (synthetic) images with parts of the target domain (real) images, and then reconstructed the style-converted synthetic images through Inverse Fast Fourier Transform (iFFT). Subsequently, these synthetic-real hybrid images and their original labels were used to train the model, reducing the differences in low-level features (e.g., texture, shape) between the two domains. Meanwhile, to further improve the model’s generalization ability on real images, we introduced an entropy minimization-based regularization term and a pseudo-label self-supervised training strategy, thereby achieving efficient cross-domain adaptation. 

The real-world experimental results are presented in Tab.~\ref{tab:2}. As can be seen from Tab.~\ref{tab:2}, even in real scenarios, our method can achieve more accurate grasping precision under different illumination conditions, demonstrating the reliability and practicality of our method. 
In addition, we compared the semantic mask generation accuracy of different multimodal methods on the Real-world dataset. As indicated in Tab.~\ref{tab:3}, compared with other methods, our method not only achieves higher mask generation accuracy but also exhibits greater stability with smaller fluctuations when facing varying illumination conditions.

Furthermore, to verify the effectiveness of our proposed domain transfer strategy, we conduct an ablation analysis, with the experimental results presented in Tab.~\ref{t222}. As can be seen from Tab.~\ref{t222}, if our transfer strategy is removed (i.e., training relies solely on real-world data), the performance of both GraspALL and DarkSeg degrades significantly. This demonstrates that our transfer strategy can effectively transfer GraspALL and all baseline models from their MIGG-trained weights to the real-world environment, and leverage the stable feature knowledge from the simulation environment to enhance the model's generalization ability and robustness in the real world.
\section{Strategy for Low-Quality Depth Maps}
\label{s4}
Our GraspALL relies heavily on the structural information of depth maps. 
However, depth maps are often affected by factors such as sensor noise, leading to unstable depth values or local missing regions. 
This makes it difficult to provide accurate geometric structural features for GraspALL. To address this, we designed a two-stage depth enhancement strategy (TSD-En). 
Unlike previous methods~\cite{R56,R54} that only perform noise smoothing, our strategy not only emphasizes noise smoothing but also focuses on resolving the problem of local depth value holes caused by noise, thereby improving the structural integrity of depth maps.

First, for the input depth map, we first adopt bilateral filtering to smooth random noise while preserving edge structures. For each pixel $p$ in the depth map, the filtered depth value $\mathbf{D}_{b}\left(p\right)$ is defined as:
\begin{equation}
\begin{array}{l}
\mathbf{D}_{b}\left(p\right)=\frac{1}{W_{p}}\sum_q \limits{\mathbf{G}_{\sigma}^{s}\left(p,q\right)\cdot\mathbf{G}_{\sigma}^{i}\left(p,q\right)\cdot q},\\
\end{array}
\label{ee4}
\end{equation}
where $q$ denotes a point within the neighborhood window $\mathcal{N}\left(p\right)$ of pixel $p$: $q\in\mathcal{N}\left(p\right)$. 
$\frac{1}{W_{p}}$ is a normalization coefficien: ${W_{p}}=\sum_q \limits{\mathbf{G}_{\sigma}^{s}\left(p,q\right)\cdot\mathbf{G}_{\sigma}^{i}\left(p,q\right)}$.
$\mathbf{G}_{\sigma}^{s}\left(\cdot\right)$ and $\mathbf{G}_{\sigma}^{i}\left(\cdot\right)$ represent the spatial and the intensity Gaussian kernel:
\begin{equation}
\begin{array}{l}
\mathbf{G}_{\sigma}^{s}\left(p,q\right)=\mathbf{exp}\left(-\frac{||p-q||^2}{{2\sigma_{s}^{2}}}\right),\\[6pt]
\mathbf{G}_{\sigma}^{i}\left(p,q\right)=\mathbf{exp}\left(-\frac{| p-q |^2}{{2\sigma_{s}^{2}}}\right).\\
\end{array}
\label{ee4}
\end{equation}

Specifically, $\mathbf{G}_{\sigma}^{s}\left(\cdot\right)$ is used to weight pixels that are closer within the neighborhood and regulate the neighborhood range, and $\mathbf{G}_{\sigma}^{i}\left(\cdot\right)$ is designed to avoid smoothing across edges. The aforementioned process can efficiently suppress random noise while maintaining critical geometric features such as garment wrinkles and edges.

After filtering, some regions may develop holes due to noise correction or sensor defects. To restore the complete depth structure, we need to further perform hole interpolation completion. For the missing pixel $p \in \Omega_{\text{hole}}$, its completed depth value $\mathbf{D}_c(p)$ is expressed as:
\begin{equation}
\begin{array}{l}
\displaystyle
\mathbf{D}_c(p)=\frac{\sum{}_{q\in\mathcal{N}_{v}\left(p\right)}|w\left(p,q\right)\cdot\mathbf{D}_{b}\left(q\right)}{\sum{}_{q\in\mathcal{N}_{v}\left(p\right)}|w\left(p,q\right)},\\
\end{array}
\label{ee4}
\end{equation}
where $\mathcal{N}_{v}\left(p\right)$ represents the valid pixel set surrounding pixel $p$. 
The weight $w\left(p,q\right)$ is defined as follow:
\begin{equation}
\begin{array}{l}
\displaystyle
w\left(p,q\right)=\mathbf{exp}\left(-\frac{\nabla|\mathbf{D}_{b}\left(q\right)|^2}{{2\sigma_{s}^{2}}}\right)\cdot\mathbf{exp}\left(-\frac{||p-q||^2}{{2\sigma_{s}^{2}}}\right),\\
\end{array}
\label{ee4}
\end{equation}
where, $\nabla|\mathbf{D}_{b}\left(q\right)|^2$ denotes the spatial gradient of the depth map at pixel $q$, which is used to describe the local intensity of depth changes. 
The introduction of the gradient term can effectively prevent incorrect interpolation across depth edges, making the completion process more consistent with the real geometric structure of the object. 
\begin{table}[t]
  \centering
  \vspace{0.1cm}
  \scalebox{0.8}{
    \begin{tabular}{p{9.5em}|p{5.8em}<{\centering}|p{4.5em}<{\centering}|p{4.5em}<{\centering}}
    \toprule
    \toprule
 \rowcolor[rgb]{ .8,  .8,  .8} \textbf{Method}& \textbf{Luminance}   & \textbf{mIoU}  & \textbf{mGSR} \\
    \midrule
    GraspALL  & \textbf{0 - 20} & 82.2\% & 82.5\% \\
    \rowcolor[rgb]{ .9,  .9,  .9} GraspALL \textbf{+} ~\cite{R54} & \textbf{0 - 20} & {82.9\%}  & {83.3\%} \\
    \midrule
        \rowcolor[rgb]{ .8,  .93,  1} GraspALL \textbf{+ TSD-En} & \textbf{0 - 20} & \textbf{84.5\%}  & \textbf{86.6\%} \\
    \midrule
    GraspALL & \textbf{20 - 40}  & 83.1\%  & 84.1\% \\
    \rowcolor[rgb]{ .9,  .9,  .9} GraspALL \textbf{+} ~\cite{R54} & \textbf{20 - 40} & {83.6\%} &{85.0\%}\\
    \midrule
        \rowcolor[rgb]{ .8,  .93,  1} GraspALL \textbf{+ TSD-En} & \textbf{20 - 40} & \textbf{85.2\%} & \textbf{87.5\%}\\
    \bottomrule
    \bottomrule
    \end{tabular}%
    }
    \vspace{-0.25cm}
  \caption{The improvement brought by the TSD-En to GraspALL.}
  \vspace{-0.3cm}
  \label{tab:4}%
\end{table}%

The weight $w\left(p,q\right)$ to impose dual constraints on spatial distance and gradient intensity, ensuring the interpolation result is both continuous and non-crossing of edges, thereby preserving the object’s geometric contour.

To verify the our strategy, we conduct corresponding ablation analysis on GraspALL. 
The experimental results are shown in Tab.~\ref{tab:4}. 
As can be seen from Tab.~\ref{tab:4}, when our TSN-En strategy is adopted, both the semantic mask generation accuracy and grasp success rate of the model are improved to a certain extent under different illumination conditions. 
In addition, compared with methods that only focus on smoothing depth map noise, our strategy also achieves better performance, increasing the grasp accuracy and semantic mask generation accuracy by 5 and 3 percentage points, respectively. 
This indicates that for the processing of low-quality depth maps, it is not only necessary to smooth the abnormalities caused by noise but also to further fill the depth gaps that may be generated after noise smoothing.

\begin{figure}[t]
  \centering
  \includegraphics[height=4.15cm]{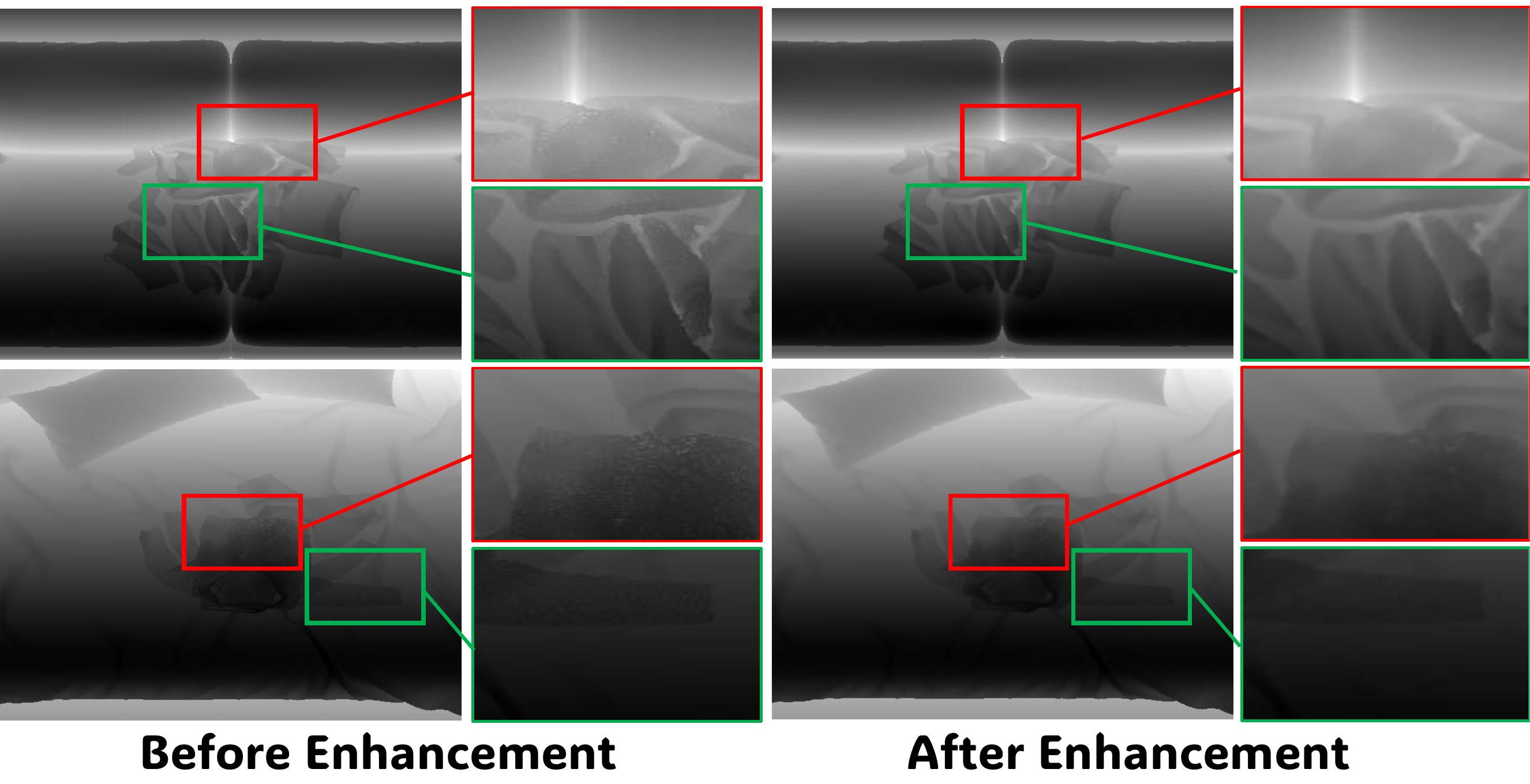}
  \vspace{-0.2cm}
  \caption{
Visualization results with and without our depth map enhancement strategy (TSD-En).
  }
  \vspace{-0.4cm}
  \label{fig:4}
\end{figure}

To intuitively demonstrate the effectiveness of our depth map enhancement strategy, we show the impact of using and not using TSD-En on depth maps, as illustrated in Fig.~\ref{fig:4}. As can be seen from Fig.~\ref{fig:4}, when depth maps suffer from noise or depth value holes due to sensor noise, our method can effectively suppress noise and fill the holes.
The above results prove that when the depth sensor generates low-quality depth maps, the model can further correct the disturbed depth information through the proposed depth map enhancement strategy, thereby providing more accurate structural information for the grasping model.
\section{Analysis of the Proposed Grasping Strategy}
\label{s6}
\begin{table}[t]
  \centering
  \vspace{0.1cm}
  \scalebox{0.8}{
    \begin{tabular}{p{7.5em}|p{6.4em}<{\centering}|p{6.4em}<{\centering}|p{3.9em}<{\centering}}
    \toprule
    \toprule
 \rowcolor[rgb]{ .8,  .8,  .8} \textbf{Method}& \textbf{Luminance$<$20}   & \textbf{Luminance$<$40}  & \textbf{Average} \\
    \midrule
    GraspALL + \cite{R8} & 17/25 (68\%)    & 18/25 (72\%)   & 70\% \\
    \rowcolor[rgb]{ .9,  .9,  .9} GraspALL + \cite{R9} & 17/25 (68\%)    & 19/25 (76\%)    & 72\% \\
    GraspALL + \cite{R7} & 19/25 (76\%)    & 20/25 (80\%)   & 78\% \\
        \midrule
    \rowcolor[rgb]{ .8,  .93,  1} GraspALL + \textbf{Ours} & \textbf{21/25 (84\%)}  &  \textbf{22/25 (88\%)} & \textbf{86\%}\\
    \bottomrule
    \bottomrule
    \end{tabular}%
    }
    \vspace{-0.2cm}
  \caption{Grasping performance of different grasping strategies.}
  \vspace{-0.15cm}
  \label{tab:6}%
\end{table}%
\begin{table}[t]
    \centering
    \scalebox{0.85}{
    \begin{tabular}{cccc}
        \toprule
        \toprule
        \textbf{Luminance} & \textbf{DarkSeg} & \textbf{BiFCNet} & \textbf{GraspALL} \\
        
        \arrayrulecolor{gray!40}\hline\arrayrulecolor{black}
        
        \rowcolor{gray!12}
        $0 - 20$ & 59.2\% & 46.7\% & \textbf{70.5\%} \\
        
        $20 - 40$ & 61.6\% & 54.5\% & \textbf{74.4\%} \\
        \bottomrule
        \bottomrule
    \end{tabular}}
    \vspace{-0.2cm}
    \caption{Analysis for our Grasping Strategy (in Simulation)}
    \label{t100}
    \vspace{-0.2cm}
\end{table}
\begin{figure}[t]
  \centering
  \includegraphics[height=5.6cm]{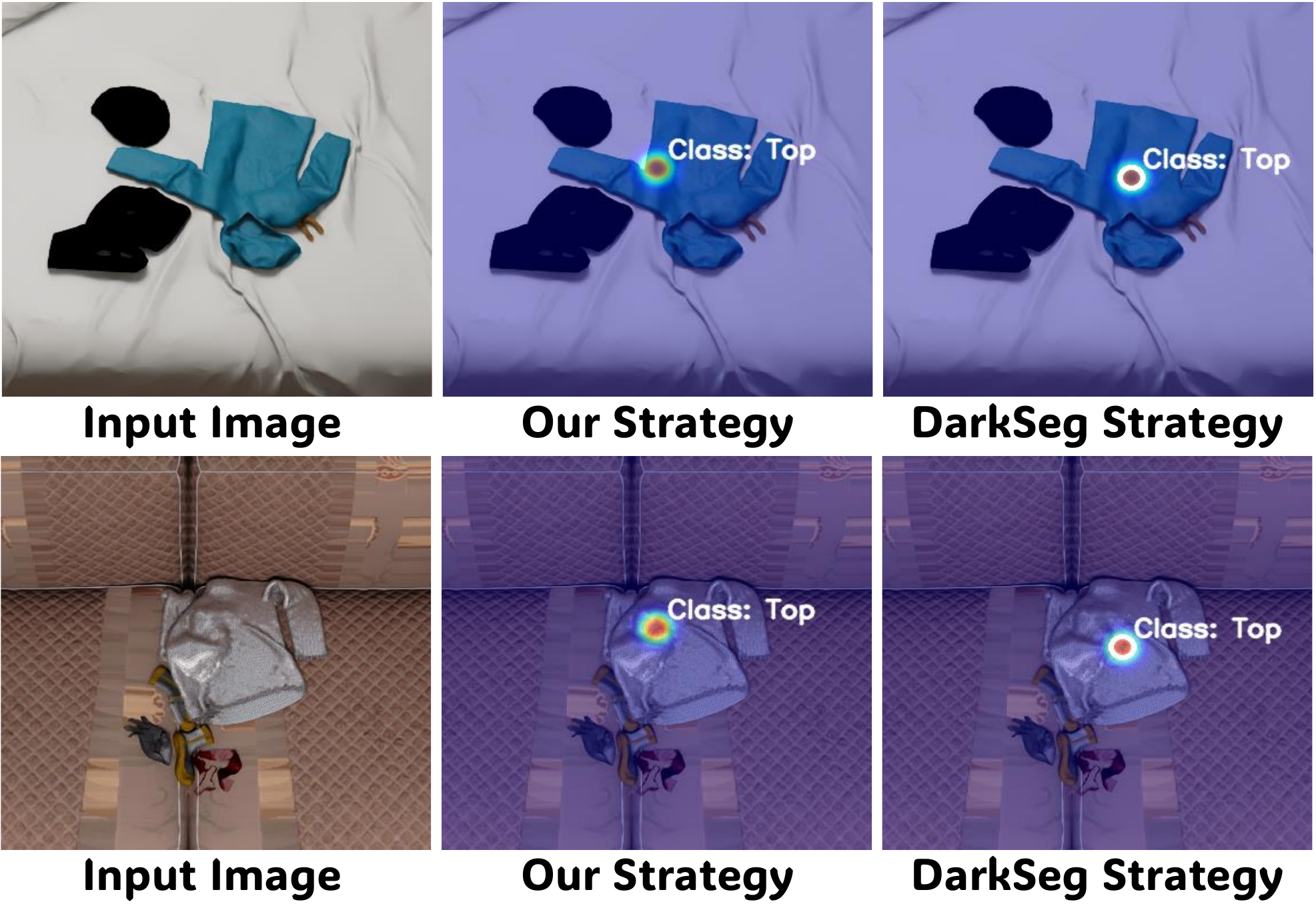}
  \vspace{-0.1cm}
  \caption{
Comparison between our grasping strategy and mainstream strategies that take the geometric center as the grasp point.
  }
  \vspace{-0.4cm}
  \label{fig:6}
\end{figure}
To obtain grasp points based on garment semantic masks, existing methods usually determine grasp points based on the geometric center of the garment or positions adjacent to the center point, aiming to avoid grasping the edge positions of the garment which may lead to dragging problems during the grasping process. 
However, the deformability of garments means that the center position is not necessarily a point with obvious wrinkles and easy to grasp, which can result in slipping during grasping and moving. 
\begin{figure*}[t]
  \centering
  \includegraphics[height=4.16cm]{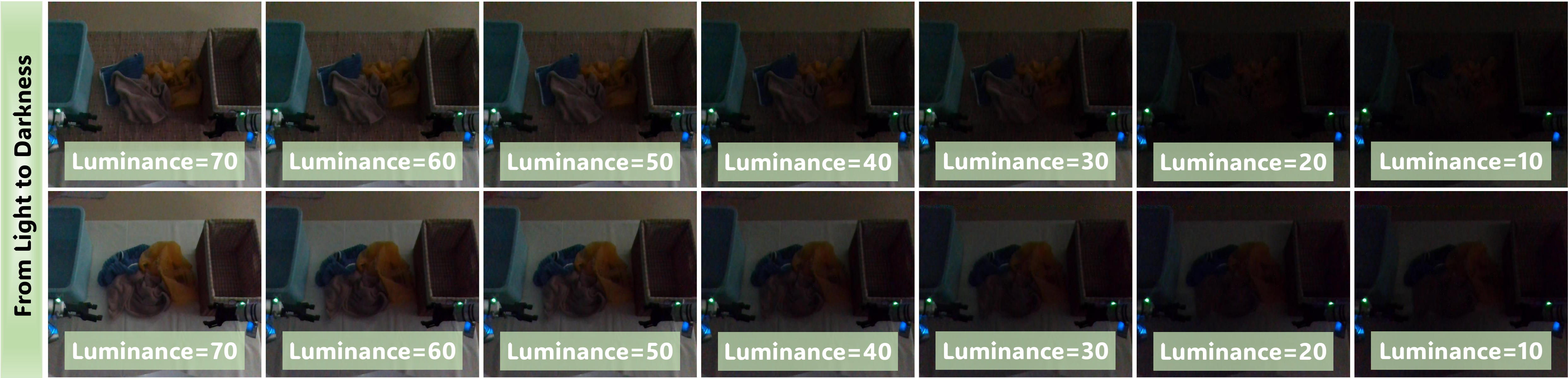}
  \vspace{-0.25cm}
  \caption{
Non-garment images (towels and shopping bags) captured under different illumination conditions in real-world scenarios.
  }
  \vspace{-0.4cm}
  \label{fig:5}
\end{figure*}
In response to this, we propose a depth optimal search strategy to find positions that are as easy to stably grasp as possible on the premise of avoiding garment dragging. To verify the effectiveness of the proposed strategy, we compare its performance with several grasp strategies that take the garment center as the grasp point. 
The results are shown in Tab.~\ref{tab:6}. 
It can be seen from Tab.~\ref{tab:6} that using our strategy can improve the grasping accuracy by 7–11\%, indicating that our strategy enables the model to achieve better grasping performance. The above results verify the correctness of our grasp strategy: when determining grasp points based on garment semantic mask regions, we should first identify candidate grasp points with obvious wrinkles from a global perspective, and then select the optimal grasp point from the candidate points according to the center point of the garment.

To more intuitively verify the performance of our method, we compare the performance differences between the proposed method and the method that takes the geometric center as the grasp point. As shown in Fig.~\ref{fig:6}, although DarkSeg—with the central point as its main grasping region—can identify the geometric center of the garment, the wrinkles at the geometric center are very weak. In contrast, the grasp points determined by our method can not only be as close as possible to the geometric center but also better locate regions with prominent wrinkles.

Moreover, in Tab.~\ref{t100}, to isolate the contribution of model perception, we fix the grasping strategy to the geometric center of the perception mask for the garment, without using depth or top-k heuristics. GraspALL still outperforms baselines, indicating that baselines are less robust to illumination changes and thus produce more class errors, while GraspALL can better adapt to illumination changes.
\section{Analysis for the Parameter $\alpha$ of the EMA }
\label{s7}
\begin{table}[t]
  \centering
  \scalebox{0.78}{
    \begin{tabular}{p{4.6em}<{\centering}|p{3.43em}<{\centering}|p{3.43em}<{\centering}|p{3.43em}<{\centering}|p{3.43em}<{\centering}|p{3.43em}<{\centering}}
    \toprule
    \toprule
 \rowcolor[rgb]{ .8,  .8,  .8} {Metrics}& {$\alpha$}=0.01 & {$\alpha$}=0.03 & \textbf{$\alpha$=0.05}  & {$\alpha$}=0.07 & {$\alpha$}=0.09 \\
    \midrule
    \textbf{mIoU} & 78.2\%  & 80.9\% & \textbf{82.8\%} & 81.1\% & 79.6\% \\
    \rowcolor[rgb]{ .9,  .9,  .9} \textbf{mGSR} & {78.3\%}  & {81.6\%} & \textbf{84.2\%} & 82.5\% & 81.6\%\\
    \bottomrule
    \bottomrule
    \end{tabular}%
    }
    \vspace{-0.1cm}
  \caption{The effect of different $\alpha$ on model (Luminance:0-40).}
  \vspace{-0.3cm}
  \label{tab:7}%
\end{table}%
In GraspALL, we adopt the EMA (Exponential Moving Average) strategy to update the generated luminance and structural compensation features to the corresponding luminance and structural response libraries. 
For the EMA strategy, we set the update momentum \(\alpha = 0.05\). To verify the rationality of this parameter setting, we analyze the impact of different \(\alpha\) values on model performance.
\begin{table}[t]
  \centering
  \vspace{0.1cm}
  \scalebox{0.79}{
    \begin{tabular}{p{6em}<{\centering}p{5.8em}<{\centering}|p{6em}<{\centering}p{6em}<{\centering}}
    \toprule
    \toprule
 \rowcolor[rgb]{ .8,  .8,  .8} \textbf{Method}& \textbf{mIoU}   &  \textbf{Method} & \textbf{mGSR} \\
    \midrule
    SegMiF~\cite{R3}  & 64.6\% & BiFCNet~\cite{R4} & 52.4\% \\
    \rowcolor[rgb]{ .9,  .9,  .9}MRFS~\cite{R2} & 67.5\% & SAM-M~\cite{R5} & 59.2\% \\
    AMDA~\cite{R1} & 68.9\% & DarkSeg~\cite{R7}  & 63.3\% \\
    \midrule
    \rowcolor[rgb]{ .8,  .93,  1} \textbf{GraspALL} & \textbf{86.7\%} & \textbf{GraspALL} & \textbf{88.3\%}\\
    \bottomrule
    \bottomrule
    \end{tabular}%
    }
    \vspace{-0.1cm}
  \caption{Performance of different methods on other objects.}
  \vspace{-0.3cm}
  \label{tab:8}%
\end{table}%

The experimental results are shown in Tab.~\ref{tab:7}. 
As can be seen from Tab.~\ref{tab:7}, when \(\alpha\) is small (e.g., 0.01 and 0.03), the model absorbs newly generated compensation features too slowly with low update efficiency, making it difficult for the features in the two response libraries to quickly adapt to diverse illumination and structural changes. 
When \(\alpha\) is large (e.g., 0.07 and 0.09), the model is overly sensitive to the update of new features and prone to interference from noise or abnormal sample features, which reduces the stability of features in the response libraries and leads to overfitting. 
Therefore, when \(\alpha = 0.05\), the model can achieve an optimal balance between the update efficiency of the response libraries and feature stability. 
It can not only timely integrate effective compensation features to adapt to scene changes but also avoid noise interference to ensure the reliability of features in the libraries.
\section{Validation of GraspALL's Generalization}
\label{s8}
To verify the generalization performance of GraspALL on other deformable objects, we collected images of shopping bags and towels (commonly seen in household scenarios) under different illumination conditions for generalization experiment validation. 
Fig.~\ref{fig:5} shows some of the shopping bag and towel images, and their collection process is consistent with that of RealData in Sec.~\ref{s3}. 
The experimental results are presented in Tab.~\ref{tab:8}. 
As can be seen from Tab.~\ref{tab:8}, compared with other methods, even when facing non-garment deformable objects such as shopping bags and towels with the interference of illumination variations, our GraspALL can still maintain high semantic mask generation accuracy and grasp success rate. 
The above experimental results indicate that our method not only performs excellently in garment grasping tasks but also possesses strong cross-deformable object category generalization ability, which can effectively adapt to different common deformable objects and complex illumination environments in household scenarios.
\begin{table}[t]
  \centering
  \vspace{0.18cm}
  \scalebox{0.7}{
    \begin{tabular}{p{4.5em}<{\centering}||p{5.6em}<{\centering}|p{5.6em}<{\centering}|p{5.6em}<{\centering}|p{5.6em}<{\centering}}
    \toprule
    \toprule
    \rowcolor[rgb]{ .8,  .8,  .8}\textbf{Luminance}&\textbf{BiFCNet~\cite{R4}} & \textbf{SAM-M~\cite{R5}} & \textbf{DarkSeg~\cite{R7}} & \textbf{Ours} \\
    \midrule
    Lu: 00 - 40 & 37.7\% \textcolor{red}{\scriptsize~(4.8-7.5)}  & 35.5\% \textcolor{red}{\scriptsize~(3.4-5.5)} & 48.8\% \textcolor{red}{\scriptsize~(5.3-8.7)} & \textbf{80.0\%} \textcolor{red}{\scriptsize~(2.9-3.6)}\\ 
    \rowcolor[rgb]{ .9,  .9,  .9} Lu: 40 - 80 & 44.4\% \textcolor{red}{\scriptsize~(3.8-4.9)} & 40.0\% \textcolor{red}{\scriptsize~(4.5-6.4)}& 53.3\% \textcolor{red}{\scriptsize~(2.5-4.4)} & \textbf{84.4\%} \textcolor{red}{\scriptsize~(1.9-3.2)}  \\
    \bottomrule
    \bottomrule
    \end{tabular}%
  }
    \caption{Statistical analysis of different methods. The red font indicates the accuracy fluctuation across three experimental rounds.}
  \vspace{-0.5cm}
    \label{tab:9}%
\end{table}%
\section{Statistical Analysis}
\label{s9}
During the grasping tests, our setup involves 15 grasping attempts per test, with a successful grasp defined as picking up the garment of the target category and placing it into the corresponding category-specific basket. 
Considering the experimental randomness, we conduct multiple repeated experiments to ensure statistical significance of the results. 
Specifically, we perform three consecutive rounds of grasping tests for each method in the real world, with 15 attempts per round, and finally compared the average grasp success rate and standard fluctuations across the three rounds. 
As shown in Tab.~\ref{tab:9}, our GraspALL achieves the highest average grasp success rate among all methods, and the fluctuation in success rate is smaller than that of the comparative methods. 
The above results indicate that our method has obvious advantages in terms of result stability, and further proves that the experimental results are not affected by random factors and have reliable statistical significance.
\begin{figure}[t]
  \centering
  \includegraphics[height=4.23cm]{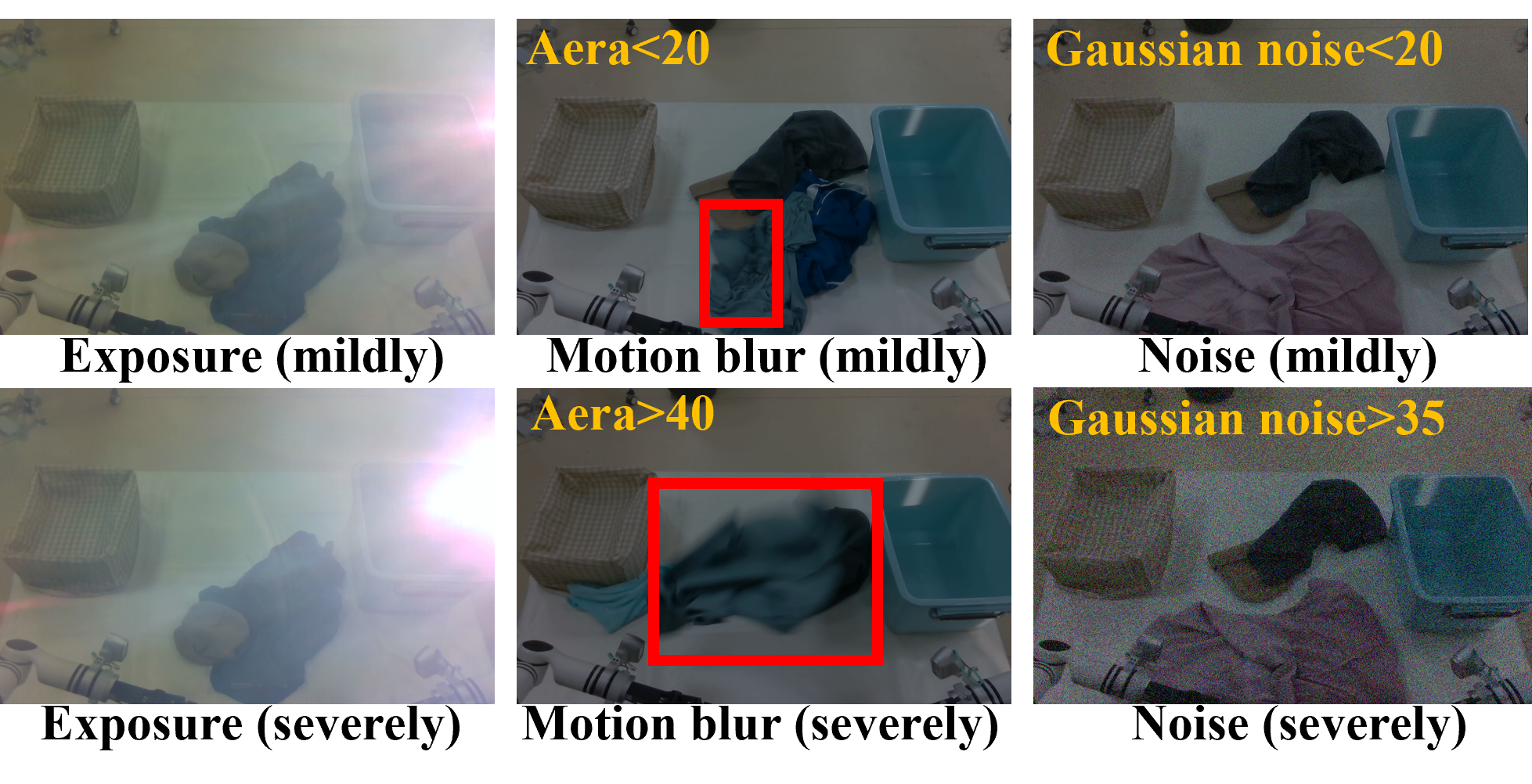}
  \vspace{-0.7cm}
  \caption{
The test samples we collected for different types of image degradation (exposure, noise, and motion blur).
  }
  \vspace{-0.2cm}
  \label{fig:111}
\end{figure}
\begin{table}[t]
    \centering
    \scalebox{0.76}{
    \begin{tabular}{cc|c|cc|cc}
        \toprule
        \multicolumn{2}{c|}{\textbf{No Corruption}} & & \multicolumn{2}{c|}{\textbf{Mildly}} & \multicolumn{2}{c}{\textbf{Severely}} \\
        \midrule
        \cellcolor{gray!15}\textbf{mGSR} & \cellcolor{gray!15}\textbf{mIoU} & 
        & \cellcolor{gray!15}\textbf{mGSR} & \cellcolor{gray!15}\textbf{mIoU} & 
        \cellcolor{gray!15}\textbf{mGSR} & \cellcolor{gray!15}\textbf{mIoU} \\
        \midrule
        \multirow{3}{*}{\textbf{80\%}} & \multirow{3}{*}{\textbf{73.4\%}} & 
        \textbf{Exposure} & 70\% & 69.7\% & 70\% & 64.3\% \\
        
        & & \cellcolor{gray!15}\textbf{Motion blur} & \cellcolor{gray!15}60\% & \cellcolor{gray!15}67.2\% & \cellcolor{gray!15}40\% & \cellcolor{gray!15}53.8\% \\
        
        & & \textbf{Noise} & 70\% & 70.8\% & 50\% & 49.5\% \\
        \bottomrule
    \end{tabular}}
    \vspace{-0.25cm}
     \caption{Validation of GraspALL Robustness.}
         \vspace{-0.5cm}
         \label{t99}
\end{table}
\section{Robustness analysis of our GraspALL}
\label{s11}
GraspALL exhibits a certain degree of robustness to real-world exposure, noise and motion blur. 
First, since the PLC Curve ID is derived from 256 RGB histogram points, even under mild corruption that degrades some details, sufficient features remain for stable matching. 
In addition, our transfer strategy leverages stable priors learned in simulation to partially compensate for corrupted RGB features. 
For validation, in Fig.~\ref{fig:111}, we collect mildly and severely corrupted samples (10 images each) for exposure, noise, and motion blur. 
As shown in Tab.~\ref{t99}, compared to no corruption, GraspALL’s performance changes slightly under mild corruption, but drops noticeably under severe corruption. 
We attribute this potential drawback to severe corruption disrupting too many effective RGB features, making PLC matching and the cross-attention unreliable. Overall, our GraspALL has certain robustness in dealing with image quality degradation problems such as motion blur. 
\end{document}